\pdfoutput=1
\documentclass[a4paper,english,11pt]{article}
\usepackage[utf8]{inputenc}
\usepackage[T2A]{fontenc}
\usepackage{graphicx,hyperref}%
\usepackage{multirow}%
\usepackage{amsmath,amssymb,amsfonts,amsthm,mathrsfs}%
\usepackage{booktabs}%
\usepackage{tikz,pgfplots,multirow,lscape,multicol,longtable}
\usepackage{microtype,xurl,authblk,libertine}
\usepackage[right=20mm,left=25mm,top=15mm,bottom=20mm]{geometry}

\usepackage[main=english,russian]{babel}
\usepackage{paralist,caption,natbib}
\usepackage{CJKutf8}
\usepackage{tipa}

\pgfplotsset{compat=1.18}

\newcommand{\hnz}[1]{\begin{CJK*}{UTF8}{bsmi}#1%
\end{CJK*}}

\newcommand{\dng}[1]{{\foreignlanguage{russian}{\textit{#1%
}}}}
\newcommand{\gt}[1]{\texttt{<#1>}}

\title{The Morphological Core of Dungan: \\A Two-Dialect Finite-State Model and a Multi-Genre Evaluation}

\author[1,2,3]{Anton~M.~Alekseev}
\author[1,2]{Sergey~I.~Nikolenko}
\affil[1]{St.~Petersburg Department of the Steklov Mathematical Institute, RAS}
\affil[2]{St.~Petersburg State University}
\affil[3]{Kyrgyz State Technical University named after I.~Razzakov}
\date{}

\begin{document}
\maketitle

\begin{abstract}
Dungan, a Sinitic language of Central Asia written in a Cyrillic-based script, is described in detail in the grammatical literature, yet the quantitative properties of its morphology in actual usage{, to the best of our knowledge,} have never  been measured systematically. This paper uses a finite-state morphological analyzer as a measuring instrument. Implemented with HFST and covering both dialect groups~--- the Gansu variety (the literary standard) and the Shaanxi variety~--- the model offers no new grammatical description; it formalizes the knowledge accumulated in Dungan studies and makes it measurable on corpora of three genres. Three results follow.
{Overt inflection is rare and limited: only 9.3\% of recognized tokens in the encyclopaedic register have an overt marker, the system has just ten categories, and degree marking is almost absent.}
Ambiguity is genuine but sharply localized: {78.1\%} of tokens receive a single analysis, and the residue sits almost entirely on two clitics, \dng{-ди} (genitive\,/\,progressive) and \dng{-ни} (locative\,/\,prospective). And the grammatical core proves effectively closed, the claim the instrument is really needed for: {between 78\% and 95\% of the tokens the analyzer fails on, depending on register, are simply absent from the lexicon, and the phenomena the model deliberately declines to implement account for at most 4.5\% of those failures.} Held-out coverage (80--85\%) is no lower than development coverage (73\%), while a stem list with no morphology already reaches 67.4\%, so the morphology is worth 5.2 points. The open frontier of Dungan is lexical. The analyzer, its sources and every evaluation script are released openly.
\end{abstract}

\section{Introduction}
\label{sec:intro}

Dungan is a Sinitic language whose speakers~--- descendants of Hui (Chinese-speaking Muslim) migrants who moved to Central Asia in the second half of the 19th century~--- live mainly in Kazakhstan, Kyrgyzstan and Uzbekistan \citep{zavyalova1996,rimskykorsakoffdyer1979}. Unlike other Chinese languages it is written in a Cyrillic-based orthography, adopted in 1953 in place of a Latin script that had itself replaced, in 1928--1932, the Arabic-script \emph{xiaoerjing} tradition \citep{zavyalova1996}. Grammatically Dungan is strongly isolating: inflection is minimal, grammatical meanings are carried mostly by function words and a small set of clitics, and the written norm does not mark lexical tone, which gives rise to extensive homography.

Although Dungan is well described in the grammatical literature, {we have been unable to locate} an open computational \emph{morphological processor} for it: a check of OLAC, Glottolog, Universal Dependencies, the Apertium and GiellaLT repositories, and a GitHub code search for \texttt{lexc}/\texttt{twol} sources tagged \texttt{dng}, finds neither an analyzer nor a treebank{, and every negative claim of this kind below is bounded by what those searches returned}.\footnote{A Dungan text corpus of roughly 600{,}000 tokens is reported to exist at Northwest Normal University (Lanzhou, PRC); we were unable to {download it and} verify its size or access conditions independently and note it only for completeness.} Dungan is not, however, absent from computational resources, and it would be wrong to imply otherwise: the Cr\'ubad\'an web crawl \citep{scannell2007} supplies a word-form frequency list with source URLs; English Wiktionary holds some 5{,}400 Dungan lemma pages carrying part-of-speech, tone and hanzi correspondence (a source this work itself draws on); Salmi's \emph{Dungan--English Dictionary} \citep{salmi2018} is an electronic lexicographic resource; and the code \texttt{dng} appears in recent massively multilingual corpora and language-identification models. What none of these provides is a grammar: a system that segments a word form, assigns it grammatical categories and generates it back. That is the gap this work fills, and precisely for less-resourced languages such a system is the basic building block for lemmatization, corpus annotation, spell checking, machine translation, and lexicography.

By origin and basic lexicon Dungan belongs to the Central Plains group of Mandarin (\hnz{中原官话}; the Guanzhong and southern Gansu varieties) and is close to Standard Mandarin \citep{zavyalova1996,salmi2023}. {What matters computationally is not genealogy but the script.} Lexicon also shows Turkic, Iranian and Slavic loans, typical of contact but less important for morphological analysis. {Chinese characters suppress written homonymy, e.g. \hnz{飯} `food' and \hnz{犯} `to violate' are graphically distinct though homophonous, while the phonographic Cyrillic of Dungan exposes it, especially because it does not record tone, the distinctive feature of any Sinitic phonological system. A large share of lexical stems are homographic due to that, and without dedicated means an analyzer cannot tell them apart.}

Inflection in the traditional sense is practically absent: no declension, no conjugation, no agreement. Grammatical meanings are expressed by word order and by particles and clitics that behave in many respects like independent words but attach phonetically to a content word. A finite-state rather than paradigm-based approach is therefore natural: the task reduces to describing the attachment of a limited marker set to stems and resolving the resulting homography \citep{koskenniemi1983}.

The aim of this work is not to give a new description of Dungan grammar (the language is described in detail) but to \emph{measure} the properties of its morphological core in actual usage: how compact and productive the inventory of inflectional categories is, what the inflectional load is and whether its profile is stable across genres, and, crucially, how \emph{closed} the core is, i.e.\ whether a finite set of markers exhausts the morphology of real text, leaving only the lexical frontier open. The instrument is a finite-state analyzer and generator covering both major dialect groups: analysis maps a word form to a lemma with grammatical tags (\dng{жынму} $\to$ \dng{жын}\gt{n}\gt{an}\gt{t1}\gt{pl}, ``lemma \dng{жын} `person', noun, animate, tone~I, plural''), generation is the inverse mapping. The analyzer is implemented with HFST \citep{linden2011} and accompanies this paper as supplementary material.\footnote{The complete analyzer~--- lexc/twol sources, both dialect lexicons, the gold test suite, the CI configuration and every evaluation script that produces a number reported here~--- is available at \url{https://github.com/alexeyev/dungan-finite-state-morphology}.}

A principled commitment of this work is that the model does not claim a new grammatical description of Dungan but converts into computable form the knowledge already accumulated in Dungan studies. Every formalized phenomenon is traced, wherever possible, to an existing description~--- plural, genitive, aspect markers, locative, classifiers all rest on published accounts \citep{dragunov1940,kalimov1968,imazov1982,zevakhina1997}, checked case by case (Section~\ref{sec:morphology}, Appendix~\ref{app:decisions})~--- and decisions with no description in the accessible literature are explicitly flagged as provisional. {The analyzer is released accordingly: as a reproducible resource open to verification, correction and extension by Dungan speakers and specialists.}

Our contributions are: a two-dialect finite-state model of Dungan morphology grounded in the existing descriptions, released with tests and CI; {what is, to our knowledge, the first} \emph{quantitative profile} of inflection in usage~--- frequency, productivity and load across three genres (Section~\ref{sec:profile})~--- reported with the sensitivity of each figure to how the metric is defined, rather than as a point estimate; a quantification of \emph{ambiguity} and of where it sits ({78.1\%} of recognized forms are unambiguous; the residue is concentrated on the clitics \dng{-ди} and \dng{-ни}); and a direct test of \emph{core completeness} (Section~\ref{sec:closure}) that classifies every recognition failure as lexical or morphological instead of inferring closure from a coverage figure~--- {at most 4.5\% of failures are attributable to phenomena the model deliberately omits}, against a no-morphology baseline that the model beats by 5.2 points.

\section{Related Work}
\label{sec:related}

The finite-state approach goes back to Koskenniemi's two-level model \citep{koskenniemi1983}, in which one description of the lexical--surface relation works in both directions; \citet{kaplan1994} showed that both ordered rewrite rules and two-level grammars are regular relations realisable by transducers, which is what makes analysis and generation invertible in a single mechanism. The practical toolkit is presented by \citet{beesley2003} and surveyed by \citet{karttunen2005}. The two largest platforms of this kind, Apertium \citep{khanna2021} (machine translation between closely related languages) and GiellaLT \citep{moshagen2013} (Uralic and Siberian areas), provide the infrastructure for building and testing such analyzers; the present work follows their conventions (tag format, lexc/twol organization, coverage and precision testing \citealp{washington2012}). Recent analyzers in this tradition include those for Kurmanji and Central Kurdish \citep{ahmadihassani2020,naserzade2023} and Maithili \citep{rahi2020}. {Finite-state lexica have also been used to recover a suprasegmental the script omits, in both standard cases for stronger reasons than ours: \citet{meurer2011} incorporates Abkhaz stress because stress position governs the surface realization of {[\textschwa]} (schwa), so the transducer cannot parse orthographic forms without it, and \citet{reynolds2016} generates Russian stress for language learners, where Constraint Grammar disambiguation over the transducer raises accuracy  \citep{bick2015}. Our tone tag is lexical, never touching the surface; we return to the difference in Section~\ref{sec:morphology}.}

Dungan grammar is described in detail in a body of (mostly Russian-language) scholarship. The first scientific description is by \citet{dragunow1936} and \citet{dragunov1937}, who characterized Dungan as an \emph{independent} language, distinct from the other known Chinese languages, with the Gansu dialect as its leading, literary, variety.\footnote{``\dots das Dunganische\dots als eine selbst\"andige Sprache zu betrachten, die sich von allen \"ubrigen uns bekannten chinesischen Sprachen unterscheidet''; ``In dialektischer Hinsicht ist es die Gansu-Mundart, die bereits zu einer Literatursprache wird'' \citep[S.~35]{dragunow1936}. As the authors note, at the time of writing Dungan had not yet been scientifically studied (``\dots da das Dunganische bis jetzt noch nicht wissenschaftlich untersucht worden ist'').} 
{The same first decade of research produced Polivanov's contributions:
alongside his part in the 1928--1932 Latinization of the script, he wrote
the first school grammars of the emerging literary language (Frunze,
1935--1936) and two studies in the 1937 Frunze orthography volume~--- the
phonological system of the Gansu dialect \citep{polivanov1937phon} and the
tonal analysis this model draws on throughout \citep{polivanov1937}.}
\citet{dragunov1940} then first described the categories of aspect and tense, followed by Kalimov \citep{kalimov1958,kalimov1968}, Imazov \citep{imazov1982,imazov1987}, Zavyalova \citep{zavyalova1979,zavyalova1996} and Salmi \citep{salmi1984,salmi2018}; a summarising encyclopaedic description is \citet{zevakhina1997}, and individual categories, in particular the adjective, have been studied in fieldwork \citep{zevakhina2001}. Among recent work, \citet{honkasalo2024} {presented} a contact-linguistic description of spoken Kazakhstani Gansu Dungan from 2022--2023 field recordings, which documents deep Russian influence and is a reminder that written (literary) Dungan, on which the present model rests, and the living spoken varieties differ considerably. 
{We have likewise been unable to locate computational resources carrying grammatical annotation; the electronic resources that do exist (Section~\ref{sec:intro}) supply word lists and lexicography rather than grammar; for languages in a comparable position the standard route to such a corpus is
precisely a rule-based analyzer feeding a corpus platform such as
Tsakorpus \citep{arkhangelskiy2019}, which is the downstream use the
present analyzer is built for.}

The present work adds {almost} nothing to this corpus of descriptions but builds on it, translating established knowledge into a reproducible computational form and \emph{measuring} it in usage, filling what remained outside its predecessors' scope only as explicitly flagged provisional decisions.

\section{Language Material and Sources}
\label{sec:data}

The main lexical source is Yanshansin's \emph{Concise Dungan--Russian Dictionary} \citep{yanshansin2009}, openly available as a PDF: about twelve thousand entries, each with a tone mark (Roman numerals I--III) and a Russian gloss; 6{,}987 of them are mined into the analyzer's lexicon.{\footnote{The lexicon is not exclusively dictionary-derived, and the remainder is small but not negligible. Beside the 6{,}987 Yanshansin entries the stem files hold 288 lemmas harvested from English Wiktionary's Dungan categories (CC~BY-SA; see the Ethics Statement), 159 forms from Salmi's grammatical material and 47 from his dictionary, 145 proper names, 210 shared function words, 42 recovered adjective readings (see Limitations) and 23 dialect and core-vocabulary entries.}} Grammatical information, especially the aspect markers and counting suffixes, draws on the Soviet--Russian descriptive tradition \citep{dragunov1940,kalimov1968,imazov1982} and on Salmi's work on aspect and the lexicon \citep{salmi1984,salmi2018}. For quality evaluation and for recovering individual constructions we used a small trilingual corpus (Dungan text with Chinese-character and Russian versions){, from which we also built the correctness control set of Section~\ref{sec:eval}}; the character correspondence is key~--- since Cyrillic does not record tone, it is the character that identifies unambiguously which morpheme underlies a word form. The bulk of the development corpus is 126 articles from the Dungan Wikipedia incubator, 12{,}337 tokens; {the parallel material adds 384 (folk narrative 156, the Shyvaza poem 47, glossed sentences 181), for a development-corpus total of 12{,}721}.

One dependency bears on Section~\ref{sec:eval}. The dictionary and both held-out texts are published by the Institute for Bible Translation, and {38} of the 126 Wikipedia articles ({15.9\%} of development tokens) are themselves religious in subject{; that classification is deliberately conservative. Biblical toponyms, ethnonyms, Roman officials and scriptural coinage are all excluded, so the figure is a lower bound}. ``Held-out'' below means \emph{not used during development}; it does not mean lexically or stylistically independent.


\section{Analyzer Architecture}
\label{sec:arch}

Dungan falls into two dialect groups, Gansu and Shaanxi, whose differences, though regular, affect the most frequent items: copula, personal pronouns, classifiers. The literary standard is the three-tone Gansu variety, because, as \citet{salmi2023} remarks crediting Yanshansin, the first teachers of the Dungan school in Frunze {(now Bishkek)} were Gansu speakers.

The design is ``shared core + thin dialect layer'': the majority of the morphology is described once, in a shared module, and dialect differences are factored into separate files defining uniquely named hook lexicons referenced from the shared root lexicon. Grammatical tags are multichar symbols following Apertium conventions: part of speech (\gt{n}, \gt{vblex}, \gt{adj}, \dots), nominal features (\gt{an}/\gt{nn}, \gt{sg}/\gt{pl}), verbal aspect (\gt{pst}, \gt{prog}, \gt{fut}, \gt{exp}), and others.

\begin{figure}[htbp]
  \centering
  \includegraphics[width=0.6\linewidth]{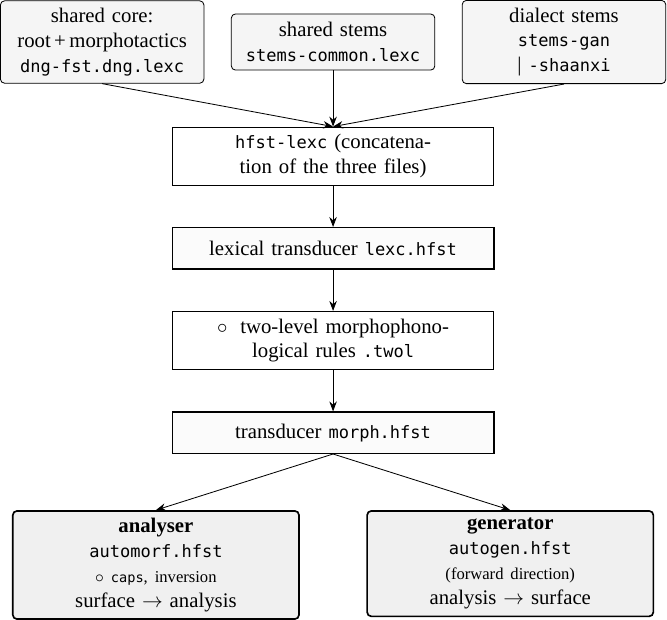}
  \caption{Build pipeline of the ``shared core + thin dialect layer'' architecture: lexical sources compile to a lexical transducer, composed with the twol rules; analyzer and generator are derived from the shared \texttt{morph}.}
  \label{fig:arch}
\end{figure}

The build is a pipeline of finite-state operations (Figure~\ref{fig:arch}): the root morphotactics, the shared stem files and one dialect stem file are concatenated and compiled by \texttt{hfst-lexc}, composed with the twol rules to yield \texttt{morph}, and from it the generator and the analyzer (by inversion, after composition with a case-folding transducer) are derived.

\begin{table}[htbp]
  \centering
  
  \normalsize
  \begin{tabular}{lll}
    \toprule
    \textbf{Meaning} & \textbf{Gansu} & \textbf{Shaanxi} \\
    \midrule
    copula & \dng{сы} & \dng{шы} \\
    `to be, become' & \dng{ви} & \dng{вый} \\
    1sg `I' & \dng{вә} & \dng{ңә} \\
    1pl `we' & \dng{вәму} & \dng{ңәму / ңай} \\
    3pl `they' & \dng{таму} & \dng{ана / тана} \\
    classifier & \dng{гә} & \dng{гуә, гы} \\
    \bottomrule
  \end{tabular}
  \caption{Regular dialect correspondences formalized in the model.}
  \label{tab:dialects}
\end{table}


The split costs little: the two analyzers differ by 1.2 points on the same corpus, and essentially the whole difference is one entry~--- \dng{вый} `to be, become', absent from the Gansu lexicon, occurs 147 times (1.16\% of the corpus). The comparison measures a lexicon gap rather than a grammatical distance.

\section{Modelled Morphology}
\label{sec:morphology}

Inflection is modest, but the system of parts-of-speech and of the bound morphemes attaching to them is well developed: \citet{salmi2023} observes that Dungan has visibly more morphological markers than {other Sinitic varieties}. \citet{imazov1979} distinguishes fourteen parts of speech; the analyzer realises eleven of them (participle, converb and preposition are not tagged: \dng{ба} \hnz{把} is a particle and \dng{лян} \hnz{連} a conjunction){, splits Imazov's single conjunction class into the two tags \gt{cnjcoo} and \gt{cnjsub},} and adds \gt{np} and \gt{cop}, so that its own inventory also numbers 14{~tags}, which is a coincidence. Tables~\ref{tab:pos} and~\ref{tab:feats} (Appendix~\ref{app:tags}) give the fourteen part-of-speech tags and the ten overtly marked inflectional categories, listing separately the five tags the analyzer emits that are lexical or zero-marked (\gt{sg}, \gt{pers}, \gt{p1}--\gt{p3}), since Section~\ref{sec:profile} turns on that distinction. {Every string shown there with tags is verbatim \texttt{hfst-lookup} output; cells giving only a lemma, a hanzi (a logogram) and a gloss identify the example word and are not analyzer output.}

A key boundary is that between inflection and word formation. {Two-syllable nouns are formed by compounding, suffixation (the ``substantive'' \dng{-зы}, \dng{-р}) and reduplication, all yielding lexicalized, idiomatic units \citep{tsunvazo1955,zevakhina2018,zevakhina2019}; adverbs are formed with multifunctional markers as well (\dng{-ди}, \dng{-ха}, \dng{-шон}, \dng{-му}, the postposition \dng{-ни}) rather than with a dedicated suffix as in Chinese, and the formation is lexicalized \citep{tsunvazo1963}.} Such words are stored as whole dictionary stems (\gt{adv}). What is modelled productively is inflection {in the strict sense:} number, case clitics, aspect, degrees of comparison.

\begin{table}[t]
  \centering
  \normalsize
  \begin{tabular}{lcp{3.45cm}p{3.65cm}}
    \toprule
    \textbf{Marker} & \textbf{Tag} & \textbf{Meaning} & \textbf{Example} \\
    \midrule
    \dng{-ли} & \gt{pst} & perfective (\hnz{哩/了}) & \dng{нянли} `read' \\
    \dng{-дини} & \gt{prog} & progressive (\hnz{底呢}) & \dng{щедини} `is writing' \\
    \dng{-ди} & \gt{prog} & stative (\hnz{底}) &~--- \\
    \dng{-ни} & \gt{fut} & prospective (\hnz{呢}) & \dng{щени} `will write' \\
    \dng{-гуә} & \gt{exp} & experiential (\hnz{過}) & \dng{нянгуә} \\
    \bottomrule
  \end{tabular}
  \caption{The five-member aspect--tense paradigm of the verb.}
  \label{tab:aspect}
\end{table}

The aspect--tense system is obligatory: a verb without one of the five markers forms not a sentence but an incomplete phrase\footnote{Sharply formulated by \citet[S.~48]{dragunow1936}; quoted in full in Appendix~\ref{app:quotes}.} \citep{dragunow1936,dragunov1940,kalimov1968,salmi1984,salmi2023}. The five markers are summarized in Table~\ref{tab:aspect}.

Every modelled phenomenon is traced to its sources with page-level anchors in Appendix~\ref{app:decisions}, which presents the model as a single decision table: \emph{phenomenon} $\to$ \emph{what the sources say} $\to$ \emph{how it is formalized}. Argumentation that does not reduce to a table row follows as notes; verbatim quotations underpinning each rule, with page and paragraph anchors, are in Appendix~\ref{app:quotes}. Because much of this literature is hard to locate, Appendix~\ref{app:resources} additionally surveys the Dungan resource landscape with locators.

\paragraph{Decisions needing argument.}
Four choices do not reduce to a table row and are set out in full in Appendix~\ref{app:extnotes}. Derivation is treated lexically, since \dng{-зы} is largely lexicalized and segmenting it would create spurious homonymy \citep{tsunvazo1955}. The tag \gt{pst} is a practical label for an aspectual marker \citep[p.~15]{dragunov1940}, as \gt{fut} is for a prospective one. Our five-way system differs from Salmi's \citeyearpar{salmi1984}: it distinguishes progressive \dng{-дини} from stative \dng{-ди} and does not include Salmi's past habitual category. And, {important for the following,} Dungan lets a marker scope over a whole phrase: an adjective predicates with aspect markers (\dng{жә-дини} `(it is) hot'; \citealp{zevakhina2001}) and in the verb--object construction the marker attaches to the object noun, \dng{ще зы-дини} `is writing' (\hnz{寫字底呢}) \citep{dragunov1940}. The noun therefore admits the aspect clitics, so the nominal genitive and the predicative progressive \emph{compete} for one surface \dng{-ди}, an ambiguity quantified in Section~\ref{sec:profile}.

\paragraph{Degrees of comparison.}
The adjective admits the comparative \dng{-щер} (\dng{да} `big' $\to$ \dng{дащер} `bigger') and the intensive-elative \dng{-дихын} (\hnz{得很} `very': \dng{да} $\to$ \dng{дадихын} `very big'); Imazov's own examples are \dng{хи} `black' $\to$ \dng{хищер} and \dng{бый} `white' $\to$ \dng{быйдихын} \citep{imazov1979,imazov1982}. We keep the released analyzer's tag name \gt{sup}, but as a label only: \citet{imazov1982} places \dng{-дихын} in the complex comparative and \citet{zevakhina2001} calls it a suffix of full predication; the analytic superlative \dng{зыю} is not modelled.

\paragraph{Invertibility and tag order.}
Analysis and generation are invertible: the tag sits on the upper side of the transducer, the surface marker on the lower, so \dng{жынму} analyses as \dng{жын}\gt{n}\gt{an}\gt{t1}\gt{pl} and that tag string generates exactly \dng{жынму}. {The tone tag belongs to that upper string rather than annotating it, so the generator requires it: \dng{жын}\gt{n}\gt{an}\gt{pl}, identical but for the missing \gt{t1}, is rejected. Inversion is exact but not one-to-one: \dng{жынму} also analyses as \dng{жынму}\gt{n}\gt{an}\gt{t1}\gt{sg}, from a plural the dictionary lexicalizes as an entry of its own, and both tag strings generate \dng{жынму} back. Invertibility guarantees that every analysis returns to the form it came from, not that the analysis is unique; Section~\ref{sec:profile} quantifies how often it is not.} Further examples, the tag-order convention, and the transducer graph of Figure~\ref{fig:fst} are in Appendix~\ref{app:extnotes}.

The markers discussed above attach to the stem in a fixed order; Table~\ref{tab:template} (Appendix~\ref{app:tags}) summarises this order for the four main classes as a morphotactic template.

\section{Morphophonology and Cyrillic Orthography}
\label{sec:morphophon}

Formalizing morphology as a finite-state machine forces explicit decisions where a prose description can remain indeterminate; they cluster at the interface of morphophonology and the Cyrillic script. 

The first, the central one is the lexical-tone decision of Section~\ref{sec:morphology}. 

The second is erization (\hnz{兒化}, final \dng{-р}), which unlike tone \emph{is} written. Here, attested erized forms are listed as separate stems. We do not model erization productively because this would also require an account of the accompanying tone alternations, which are not represented in the orthography.

The third decision is the affix--clitic boundary. The aspect and case markers (\dng{-ли}, \dng{-ди}, \dng{-ни}, \dng{-му}) sit between affix and clitic: phonetically bound to the content word, syntactically in many ways autonomous. The analyzer treats these elements as bound inflectional markers rather than as separate syntactic words. This keeps the model simple and makes the analysis directly testable on corpora. However, it does not capture uses that depend on sentence context, such as the omission of an aspect marker after preposed negation (see Limitations).

These decisions leave almost nothing for the two-level component, which is itself a result. Each dialect's \emph{twol} file has exactly two active rules, deleting the affix boundary \texttt{\%>} and the clitic boundary \texttt{\%+}. Two further constraints (the \dng{ў}/\dng{у} alternation after labials, the Shaanxi palatalization \dng{дь}/\dng{ть} $\to$ \dng{җь}/\dng{чь}) are documented but vacuous, since the lexicon stores stems in final orthography. Dungan has essentially no morphophonemic alternation at the boundaries this model draws: all descriptive work is done by the lexc morphotactics, and composition with \emph{twol} is {as of now} a formality, the expected shape for a strongly isolating language.

\section{A Morphological Profile of Usage}
\label{sec:profile}

A finite-state model can not only analyze word forms but also \emph{measure} how the category inventory of Section~\ref{sec:morphology} behaves in real text. We applied the analyzer to three corpora: an encyclopaedic corpus consisting of Dungan Wikipedia articles and parallel texts (12{,}543 tokens\footnote{Single-letter tokens, which are text-extraction artefacts, are excluded here. The resulting total is slightly lower than the development-corpus total reported in Section~\ref{sec:eval} (12{,}721 tokens), where no such filter is applied. This difference does not affect the conclusions.}); a folkloric corpus of folk proverbs (14{,}016 tokens); and an Old Testament narrative corpus consisting of the Dungan Pentateuch (119{,}920 tokens). For each corpus, we counted the inflectional tags assigned to every recognized token.

Two properties of the instrument shape every figure below. 
\begin{enumerate}[(i)]
    \item What counts as \emph{inflection}? The analyzer emits five tags with no overt exponent: the zero-marked singular \gt{sg} and the pronominal \gt{pers}, \gt{p1}--\gt{p3}. Thus a bare \dng{жын} `person' analyses as \dng{жын}\gt{n}\gt{an}\gt{t1}\gt{sg}. Counting such tokens as inflected would measure the tagset rather than the language: on the encyclopaedic register those five tags alone contribute {5.9 of the 15.2} points an unrestricted definition yields. We count a token as inflected only if it carries an overtly marked category.
    \item \emph{Which} analysis is counted? \texttt{hfst-lookup} returns all analyses at weight zero, so its first output line reflects transducer path order rather than any disambiguation; we report the first-reading figure together with the any-reading figure, which bracket the true value.
\end{enumerate}

On that definition the share of recognized tokens bearing an overt inflectional marker is 9.3\% in the encyclopaedic register (any-reading upper bound 13.7\%), {13.4\%} in the folkloric (14.2\%) and 27.1\% in the narrative (27.6\%). The inflectional \emph{load} is low and strongly genre-dependent, densest in narrative; the bulk of every text consists of unmarked stems and function words, as the isolating character of the language predicts.

\begin{table}[t]
  \centering
  \small
  \setlength{\tabcolsep}{3.1pt}  
  \begin{tabular}{lrrrrrr}
    \toprule
    & \multicolumn{2}{c}{\textbf{encycl.}} & \multicolumn{2}{c}{\textbf{folklore}} & \multicolumn{2}{c}{\textbf{narrative}} \\
    \textbf{category} & \% & lem. & \% & lem. & \% & lem. \\
    \midrule
    genitive \dng{-ди} & {2.23} & {95} & {3.29} & {161} & {4.82} & {395} \\
    perfective \dng{-ли} & 1.43 & 79 & 2.53 & 109 & 3.44 & 458 \\
    progressive \dng{-ди}/\dng{-дини} & {1.36} & {67} & {2.50} & {124} & {4.05} & {360} \\
    prospective \dng{-ни} & {0.82} & 50 & {2.96} & 153 & 3.00 & 468 \\
    locative \dng{-ни}/\dng{-шон} & 1.63 & 39 & 1.53 & 43 & 3.12 & 153 \\
    plural (anim.) \dng{-му} & 1.46 & 20 & 0.27 & 9 & 7.11 & 35 \\
    possessive \dng{-ди} (prn) & 0.66 & 6 & 0.27 & 4 & 2.77 & 8 \\
    experiential \dng{-гуә} & 0.02 & 2 & 0.09 & 8 & 0.13 & 17 \\
    elative \dng{-дихын} & {0.04} & {3} & 0.00 & 0 & 0.04 & {15} \\
    comparative \dng{-щер} & 0.01 & 1 & 0.00 & 0 & 0.00 & 1 \\
    \midrule
    \emph{overt inflect.\ load} & \multicolumn{2}{c}{9.3\%} & \multicolumn{2}{c}{{13.4\%}} & \multicolumn{2}{c}{27.1\%} \\
    \bottomrule
  \end{tabular}
  \caption{Profile of inflectional categories in three genres: frequency (\% of recognized tokens carrying the category in the first analysis) and productivity (distinct carrier lemmas). Only categories with an overt exponent are counted; the zero-marked \gt{sg} and the pronominal subclass tags are excluded (see text). Gansu analyzer.}
  \label{tab:profile}
\end{table}

Composition and \emph{productivity} complete the picture (Table~\ref{tab:profile}). The attributive-genitive \dng{-ди} leads every register on both counts, spanning {95 to 395} lemmas, though the three aspect markers taken together exceed it. The rest are stratified: the locative and animate plural are moderately frequent~--- the plural spikes to 7.1\% in the Pentateuch, where collective reference to peoples is constant, but on 35 lemmas only, frequent rather than productive~--- while the experiential and the two degree markers are \emph{vacant}, together under {two tenths of a percent} everywhere.

The rank order is reasonably stable \emph{across genres}. Over the ten categories of Table~\ref{tab:profile}, Spearman's $\rho$ is 0.88 between the encyclopaedic and narrative domains ($n=10$, one-sided permutation $p = 0.001$), 0.74 between encyclopaedic and folkloric ($p = 0.009$) and 0.70 between folkloric and narrative ($p = 0.015$). The poles stay fixed: the genitive and the aspect markers at the top, the degree markers and the experiential at the bottom. The middle of the ranking reorders freely with genre, most visibly the plural. With ten categories the numbers are illustrative; the same small inventory, with the same extremes, describes texts of widely different topic and style; the ordering itself is not genre-invariant.

Finally, \emph{ambiguity} is limited: {78.1\%} of recognized development-corpus tokens receive a single analysis, {15.9\%} two, the remaining 6.0\% three or more. Tone tags contribute little: ignoring them raises the unambiguous share only to {78.9\%}, and at the lemma level alone 98.4\% of tokens are unambiguous. The residue sits mostly on the two polyfunctional clitics. Among genuinely suffixed \dng{-ди} forms (299 tokens; particle \dng{ди} excluded), 49.5\% receive both a genitive and a progressive reading, 20.1\% genitive-only, and 30.4\% progressive-only; among suffixed \dng{-ни} forms (77 tokens), 68.8\% receive both a locative and a prospective reading and the entire unambiguous residue is prospective. These two clitics are the natural targets for contextual disambiguation, typically a Constraint Grammar layer over the transducer \citep{bick2015} in this tradition, which is future work.

\section{Core Completeness and Evaluation}
\label{sec:eval}

The analyzer is evaluated on \emph{coverage} (the share of tokens receiving at least one analysis) and on the correctness of the readings it returns. {Correctness has two sides and we keep them apart: \emph{recall} is the share of control items whose contextually correct reading is among the readings returned, \emph{precision} the share of returned readings that are correct. The distinction matters here because the transducer deliberately leaves ambiguity unresolved, so a reading that is wrong \emph{in context} may still be a legitimate reading of the string.} These are the conventional metrics for finite-state analyzers in the Apertium/GiellaLT tradition \citep{washington2012,khanna2021}. We add a {third measure, core-vocabulary completeness (Section~\ref{sec:eval}),} specific to the question this paper targets.

We measured coverage on a single corpus of real Dungan text (folk narrative, a poem by Ya.~Shyvaza, glossed examples and 126 Dungan Wikipedia articles; 12{,}721 tokens) for both dialect analyzers, making the figures directly comparable: 72.6\% for Gansu ({9{,}235}/12{,}721), 73.8\% for Shaanxi ({9{,}389}/12{,}721). Type coverage is much lower: {39.3\%} and {39.4\%}, as a Zipfian distribution against a fixed lexicon predicts. We report it because the token figure alone flatters any analyzer. On the literary subcorpus (203 tokens) coverage is 83.3\% and 84.7\%, but that sample is too small to press: the 95\% {Clopper--Pearson} interval on 83.3\% of 203 is [{77.4}, 88.1]. For comparison, the first Apertium Kyrgyz release reported 82--87\% \citep{washington2012}; the Dungan figures sit in a similar band on a much smaller lexicon.

For an isolating language a bare word list is already a strong baseline. Matching tokens against the {7{,}207} surface stem forms of the Gansu lexicon, with no continuation classes and no clitics, already recognizes 67.4\% of development-corpus tokens against the transducer's 72.6\%! The modelled morphology is worth 5.2 points (5.2 for Shaanxi too, 68.6\%~$\to$~73.8\%), which is a meaningful but modest contribution, and demonstrates how thin Dungan inflection is.

We grew the lexicon against the same corpus on which coverage is reported, so that figure may be optimistically biased. To assess generalization, we measured coverage on two Dungan texts \emph{never used} in development~--- publications of the Institute for Bible Translation:\footnote{Electronic Dungan editions (IBT): \url{ibtrussia.org/Dungan/bible}. The texts are used solely to count recognition statistics; they are not part of the model or of any released material.} a collection of Dungan folk proverbs (folk-literary register, 14{,}016 tokens) and the Dungan Pentateuch (Old Testament narrative, 119{,}920 tokens; the edition's glossary excluded). 

Token coverage is 85.2\% on the proverbs and 79.8\% on the Pentateuch, i.e. both are \emph{above} the development corpus. Coverage is thus not an artefact of fitting to the development corpus. Two qualifications limit what ``held-out'' means here. Domain difficulty differs: proverbs rest on frequent vocabulary, while the encyclopaedic text is dense with terminology and proper names. And, as Section~\ref{sec:data} notes, they are not independent of the development material: same publisher as the dictionary, and {at least 15.9\%} of development tokens religious in subject. These samples establish that the analyzer transfers to unseen text of related register, leaving transfer to Dungan at large untested. {A direct check supports the same reading, and in a stronger form than a first estimate suggested. Rebuilding the analyzer with every corpus-mined sublexicon emptied~--- 122 stems, added because they surfaced as frequent unknowns during development~--- costs 15.6 points of coverage on the development corpus (72.6\%~$\to$~57.0\%) but only 2.6 on the proverbs (85.2\%~$\to$~82.6\%) and 2.0 on the Pentateuch (79.8\%~$\to$~77.7\%). Those additions are therefore overwhelmingly specific to the corpus they were mined from, and what carries the held-out figures is the lexicon core.}

For correctness we assembled a control set of {104 annotated items over 99 distinct} word forms from Dungan text {(the trilingual parallel material of Section~\ref{sec:data}: the glossed sentences and the folk narrative)}, fixing each reference reading by its Chinese character correspondence. {The set is our own, assembled for this evaluation because no annotated Dungan corpus exists to sample from. Every item is a form attested in that material, five of them only inside a longer word form, and its reference reading is fixed from the hanzi of the parallel version (for four items, whose character that version does not supply, from the dictionary entry). It ships with the analyzer as \texttt{tests/gold-precision.txt}, one row per item with the hanzi and a gloss, so every label can be inspected; the annotation is the authors' own, AI-assisted and unreviewed by a native speaker (Ethics Statement).} {Recall is 104/104, i.e.\ for every item the reference reading is among those returned, and precision 106/127 = 83.5\%: of the 127 readings returned, a mean of 1.22 per item, 21 are not the reference reading. Both are tone-insensitive; with the tone tag required to match they become 69.2\% (72/104) and 56.7\% (72/127), and we report both pairs, since the tone tags are a central design decision and the tone-blind figures do not test them. The 95\% Clopper--Pearson intervals are [96.5, 100] and [75.8, 89.5]. Precision here is a lower bound, and diagnostic rather than damning: the reference fixes one reading per item, so a reading legitimate for the string but wrong in context counts against it, and that is what all 21 are, six of them the \dng{-ди} and \dng{-ни} clitic ambiguities quantified in Section~\ref{sec:profile}, nine lexical homographs (\dng{эр} \hnz{二} `two' beside \dng{эр} \hnz{兒} `son'), six part-of-speech splits on function words (\dng{ни} as pronoun\hnz{你} beside postposition \hnz{裏} and particle \hnz{呢}). None is a malformed analysis: what precision measures here is the residual contextual ambiguity a Constraint Grammar layer would resolve.}

{Both figures rest on a single, circular source:}{ Yanshansin's dictionary supplies both the analyzer's stems (stem, part of speech from the Russian gloss, tone, character correspondence) and the reference readings, fixed from the same character by the same operations. Agreement therefore certifies \emph{internal consistency}, not correctness: it catches implementation errors~--- a twol rule that failed to fire, a typo in a continuation class, lexicon--reference drift~--- but not errors shared by both sides, whether an inaccuracy of the dictionary itself or a linguistic decision applied uniformly to both. Only an independent reference and adjudication by an independent party~--- a native speaker or Dungan-studies expert~--- can break the loop; both remain necessary future work, and a kit for the second path ships with the analyzer: a stratified worksheet of word forms, analyses and contexts, biased towards affixed and ambiguous forms, prepared for native-speaker annotation. Coverage is free of this loop: measured on external text, it registers only the fact of recognition.}

{As a floor on basic vocabulary the analyzer recognizes a form for each of 34 concepts of the 40-item ASJP core list \citep{wichmann2016}. The check is weak: six concepts are untested, it asks only whether some analysis exists rather than whether it matches the concept, and its concept-to-spelling mapping comes from the same dictionary as the lexicon. It probes lexicon completeness and leaves the circularity above intact.}

\label{sec:closure}
Coverage counts recognitions; it cannot by itself support the claim that the grammatical core is closed, since a high coverage figure is equally consistent with a large lexicon and an incomplete marker inventory. The claim is about what the \emph{failures} are made of, so we measure that. For every unrecognized token we ask whether stripping a candidate affix leaves a form the analyzer recognizes, classifying the failure as (A) reachable by morphology the model implements~--- a morphotactic gap; (B) reachable by a documented but deferred phenomenon (past habitual \dng{-лэ}, erization \dng{-р}, derivational \dng{-зы}); (C) reduplication $XX$ with $X$ known; or (D) a purely lexical gap. (A)--(C) are upper bounds~--- a string ending in \dng{-лэ} need not contain the marker~--- which is the direction the argument needs.

Category (D)~--- a purely lexical gap~--- accounts for {94.7\%} of failures in the encyclopaedic register, {78.4\%} in the folkloric and {80.5\%} in the narrative; (A) for {4.0, 17.1 and 15.9\%} respectively, (B) for 0.8, 3.2 and 2.3\%, and (C) for {0.5, 1.3 and 1.3\%}. Between {78\% and 95\%} of all recognition failures are purely lexical, and the phenomena the model deliberately omits~--- past habitual, erization, reduplication, derivational suffixation~--- together account for at most 1.3\% of failures in the encyclopaedic register and 4.5\% in the folkloric. \emph{This} is the evidence for a closed core: the marker inventory of Section~\ref{sec:morphology} is not visibly missing anything that running text demands, and the open frontier is lexical. On the development corpus the residue is dominated by proper names and terminology; the single largest unrecognized type in the Gansu build is \dng{вый} `to be, become', the Shaanxi form discussed in Section~\ref{sec:arch}.

The measurement also shows where the model should grow next. Category (A) is far larger on held-out text ({16--17\%}) than on the development corpus ({4.0\%}), and inspection shows it dominated by one construction: the aspect clitics \dng{-ли}/\dng{-ни} on predicative adjectives, {69} tokens in the folkloric register alone (\dng{дуәли}, \dng{лоли}, \dng{дали}, \dng{лынли}). These are documented \citep{zevakhina2001}, and the model declines to generate them on the grounds that they are barely attested in the development corpus~--- a judgement the held-out text shows to be an artefact of that corpus rather than a property of the language. A typology of failures is a work-list, which is what the instrument buys beyond a percentage.

\section{Conclusion}
\label{sec:concl}

The measurement adds up to a coherent statement: \emph{the morphological core of Dungan is compact, productive in only a few categories, and effectively closed}. 
Overt inflection touches 9--27\% of recognized tokens by register, ten categories exhaust the system and two are all but unused (Section~\ref{sec:profile}); {78.1\%} of forms are unambiguous and the residue sits on two clitics. 
Closure is established not by the coverage figure but by the composition of the failures behind it (Section~\ref{sec:closure}): {78--95\%} of unrecognized tokens are lexical gaps, and everything the model omits accounts for at most 4.5\% of them; the open frontier of Dungan is lexical. The claim is modest: it is a quantification of what the descriptive literature already asserts, the value added being that each count is tied to an explicit and inspectable \emph{grammatical} decision rather than to a proxy such as a type--token ratio \citep{bentz2016}.

Methodologically, finite-state formalization proves a convenient \emph{measuring instrument} for a low-resource grammar: it forces decisions, yields testable quantitative estimates instead of qualitative ones, and converts its own gaps into a ranked work-list through the failure typology. 
Known phenomena described but not implemented are itemized in the Limitations section. 
The analyzer, source code and tests accompany this paper and will be released openly, as a basis for corpus annotation, lexicography and machine translation and a departure point for native-speaker verification.

\clearpage
\section*{Limitations}

The work rests on written sources and the descriptive literature, not on work with native speakers; every decision in the model therefore has the status of a hypothesis, checked against existing descriptions and the corpus but not verified by a speaker. This is why the analyzer is released openly with tests, designed for correction by Dungan specialists and speakers; a prepared adjudication worksheet for native-speaker annotation ships with it.

{The correctness measurement is} circular in the sense detailed in Section~\ref{sec:eval}: reference readings derive from the same dictionary that supplies the lexicon, so the reported {100\% recall and 83.5\% precision certify} internal consistency, not independently validated correctness{; with the tone tag required to match, the pair is 69.2\% and 56.7\%}. The ASJP core-vocabulary check does not escape the loop either, since the ASJP Dungan wordlist is itself compiled from a Yanshansin dictionary. Coverage figures are free of this loop, but the held-out evaluation validates coverage and its transfer only, {not correctness.}{ The domain (register) difficulty of the held-out texts differs from the development corpus, and, as Section~\ref{sec:data} already notes,} the held-out material shares a publisher and a subject domain with the development corpus and the source dictionary, so it is unseen but not independent.

The degree-marker counts are the weakest cells in Table~\ref{tab:profile} and their vacancy is partly an artefact of the lexicon rather than a fact about Dungan. Three mechanisms contribute. The source dictionary lists 29 \dng{-щер}/\dng{-дихын} forms as whole entries, 26 of them tagged as nouns, so \dng{зощер} `earlier' and \dng{щинщер} `newer' receive only a nominal reading and never reach the \gt{comp} slot. 

{Automatic part-of-speech assignment from the Russian gloss misfires on exactly the relevant class, in two ways: an entry whose gloss is a bare qualitative adjective but whose illustrative phrase is nominal was filed as a noun (\dng{хи} `black', gloss \emph{чёрный}, illustrated by \dng{хи кўзы} `black trousers'), and an adjective homographic with a numeral or a verb lost the slot to the competing reading (\dng{бый} \hnz{白} `white' against \dng{бый} \hnz{百} `100'; \dng{го} \hnz{高} `high' against \dng{го} \hnz{告} `complain'). The released analyzer carries an additive recovery layer of 42 adjective readings: the original reading is kept and an \gt{adj} reading is added beside it, so no  analysis is lost, admitted only for short, underived, gradable stems whose gloss opens with a bare Russian qualitative adjective. Both of Imazov's canonical degree examples consequently build: \dng{хищер} analyses as \dng{хи}\gt{adj}\gt{t1}\gt{comp} and \dng{быйдихын} as \dng{бый}\gt{adj}\gt{t2}\gt{sup}. The whole-entry storage of the 29 dictionary degree forms is not fixed by that layer, and the layer is itself a hand-audited patch over an automatic assignment, not a re-derivation of it.} 

Separately, \dng{-ли}/\dng{-ни} on predicative adjectives \citep{zevakhina2001} are not modelled at all, so such tokens fall outside the table entirely. We have not quantified the lexicon's overall part-of-speech error rate; doing so is necessary future work.

The profile of Section~\ref{sec:profile} is a measurement \emph{under the model}, with two specific dependencies. Category counts are taken from the first analysis \texttt{hfst-lookup} returns, and since all analyses carry weight zero that choice is an artefact of transducer path order; we bound the effect by reporting any-reading figures alongside, but we do not resolve it, and a Constraint Grammar layer would. The counts inherit the lexicon's errors: part of speech is assigned automatically from the Russian gloss, derived forms are sometimes stored as whole stems (notably the degree forms discussed in Section~\ref{sec:profile}), and we have not quantified the resulting error rate. Measured ambiguity is likewise bounded from below by lexicon incompleteness: homographs the lexicon does not contain cannot show up as ambiguity, so the {78.1\%} unambiguous figure is an upper bound on the true share.

The model covers written literary language. Spoken Kazakhstani Gansu Dungan differs considerably: recent fieldwork documents deep Russian influence and, for instance, the collapse of the classifier system to a single \dng{гә} \citep{honkasalo2024}, so performance on transcribed speech will be worse in ways the written-text evaluation does not show. 

Within the written language, a set of documented phenomena is not modelled (each flagged as deferred where it arises): the aspect clitics \dng{-ли}/\dng{-ни} on predicative adjectives \citep{zevakhina2001}, which Section~\ref{sec:closure} identifies as the largest gap on held-out text; noun reduplication in its distributive, diminutive-hypocoristic (\dng{лянлянзы} `little face') and kinship uses \citep{tsunvazored1949,zevakhina2018}, which in Northern Chinese and Jin dialects comes with tone alternations and erization \citep{zavyalova1996}; tone alternations under reduplication and compounding, and \emph{déplacement} of the expiratory stress generally; the past habitual \dng{-лэ}/\dng{-дилэ}; productive erization and derivational suffixation; participles, converbs, imperatives; and the prefixal ordinals \dng{ту-}/\dng{ди-}/\dng{чу-} \citep{imazov1982}. Negation is partial: the prohibitive \dng{бә} (\hnz{叵/嫑}) enters the model with \gt{neg}, but the full paradigm in which preposed \dng{бу}/\dng{мә}/\dng{бә} trigger regular loss of the aspect marker (\dng{лэни} `will come'~--- \dng{бу лэ}; \dng{лэли} `came'~--- \dng{мә лэ}), which have been described systematically by \citet{dragunov1940} and repeatedly confirmed \citep{tsunvazoneg1949,kalimov1968,imazov1982}, is syntactic and deliberately not treated at the morphological level. The tone tags resolve only tonal homography, not full homonymy of stems identical in tone. 

Finally, the lexicon derives from a single dictionary of the Gansu standard; its extraction from a PDF text layer, while checked, inherits any inaccuracies of the source.

\section*{Ethics Statement}

The model and tests are built exclusively on openly available published materials. The development corpus uses Dungan Wikipedia incubator text (CC~BY-SA) and short published examples{; a further 288 lemmas in the lexicon are harvested from English Wiktionary's Dungan categories, whose content is CC~BY-SA~4.0. The analyzer is released under GPLv3, which Creative Commons designates as a one-way compatible licence for BY-SA~4.0 material, so the ShareAlike condition on that portion of the lexicon is satisfied by the release as a whole.} The held-out texts of the Institute for Bible Translation are fetched from the publisher's site at evaluation time, used only to compute recognition statistics, and are not redistributed with the analyzer. One test file contains a short excerpt of a published Dungan poem used as a recognition probe; it is flagged with a rights caveat in the source documentation and can be removed without affecting the build. 

The work concerns a low-resource minority language; the analyzer is intended as an open community resource, explicitly designed for inspection and correction by Dungan speakers and specialists, and makes no claims about speakers or communities beyond the linguistic properties of published text. 

AI assistants (Anthropic's Claude) were used throughout the project, and the scope was not limited to polishing text: for auxiliary programming (extraction of the lexicon from the dictionary PDF, the build and evaluation scripts), for editing the prose, and for assembling the {104-item correctness} gold set of Section~\ref{sec:eval}, whose reference readings were assigned from the hanzi correspondences by the same assisted pipeline that built the lexicon. 

All linguistic examples, numbers and bibliographic entries were verified by the authors against the primary sources, no previously unpublished data were shared with these services, and all substantive decisions are the {authors'}.

\bibliographystyle{abbrvnat}
\bibliography{refs}

\appendix
\clearpage

\section{Tag Inventory and Morphotactic Template}
\label{app:tags}

Tables~\ref{tab:pos} and~\ref{tab:feats} give the full tag inventory of the analyzer, and Table~\ref{tab:template} the morphotactic template. 

\begin{table}[t]
  \centering
  \normalsize
  \begin{tabular}{llp{5.1cm}}
    \toprule
    \textbf{Tag} & \textbf{Part of speech} & \textbf{Example} \\
    \midrule
    \gt{n} & noun & \dng{жын} \hnz{人} `person' \\
    \gt{np} & proper noun & \dng{самария}\gt{np}\gt{gen} \\
    \gt{vblex} & verb & \dng{нян}\gt{vblex}\gt{t3}\gt{pst} `read' \\
    \gt{adj} & adjective & \dng{да} \hnz{大} `big' \\
    \gt{adv} & adverb & \dng{бу} \hnz{不} `not' \\
    \gt{num} & numeral & \dng{йи} `one' \\
    \gt{cls} & classifier & \dng{гә} \hnz{個} `piece' \\
    \gt{prn} & pronoun & \dng{вә} `I' \\
    \gt{post} & postposition & \dng{литу} `inside' \\
    \gt{part} & particle & \dng{ба} \hnz{把} \\
    \gt{cnjcoo} & coord.\ conjunction & \dng{лян} \hnz{連} `and, with' \\
    \gt{cnjsub} & subord.\ conjunction & \dng{йинцысы} `because' \\
    \gt{cop} & copula & \dng{сы} \\
    \gt{ij} & interjection & \dng{ама} (address) \\
    \bottomrule
  \end{tabular}
  \caption{Parts of speech in the model: tag, class, example{. Cells containing tags are analyzer output; cells giving only a lemma and a gloss name the example word}.}
  \label{tab:pos}
\end{table}

\begin{table}[t]
  \centering
  \small
  \begin{tabular}{p{5.5cm}p{3.95cm}p{5.1cm}}
    \toprule
    \textbf{Tag} & \textbf{Meaning} & \textbf{Example} \\
    \midrule
    \multicolumn{3}{l}{\emph{overt inflection (counted in Section~\ref{sec:profile})}} \\
    \gt{pl} & plural (animate) & \dng{жынму} `people' \\
    \gt{gen} & genitive / attributive & \dng{самарияди} `of Samaria' \\
    \gt{loc} & locative & \dng{дащүәни} `at the university' \\
    \gt{px} & possessive (pronoun) & \dng{вәмуди} `our' \\
    \gt{pst} & perfective (demarcational) & \dng{нянли} `read (pfv)' \\
    \gt{prog} & progressive-durative & \dng{щедини} `is writing' \\
    \gt{fut} & prospective-future & \dng{щени} `will write' \\
    \gt{exp} & experiential & \dng{нянгуә} `has read (repeatedly)' \\
    \gt{comp}/\gt{sup} & comparative / elative & \dng{дащер}, \dng{дадихын} \\
    \midrule
    \multicolumn{3}{l}{\emph{lexical and zero-marked (not counted)}} \\
    \gt{an}/\gt{nn} & animacy / inanimacy & \dng{жын}\gt{an}, \dng{вынхуа}\gt{nn} \\
    \gt{sg} & singular (zero-marked) & \dng{жын}\gt{sg} \\
    \gt{t1}/\gt{t2}/\gt{t3} & lexical tone & \dng{задё} `suck out'/`fence off'/`blow up' \\
    \gt{neg} & prohibitive negation & \dng{бә} \\
    \gt{pers}, {\gt{p1}--\gt{p3}} & pronoun person & \dng{вә}\gt{pers}\gt{p1} \\
    \gt{incl} & inclusive (pronoun) & \dng{заму} `we (incl.)' \\
    \gt{dem}/\gt{itg} & demonstr.\ / interrog. & \dng{нагә} `that', \dng{са} `what' \\
    \bottomrule
  \end{tabular}
  \caption{Features assigned on top of the part-of-speech tag, split by whether they correspond to an overt exponent. Only the upper group is counted as inflection in Section~\ref{sec:profile}. {Cells containing tags are analyzer output; cells giving only a lemma and a gloss name the example word}.}
  \label{tab:feats}
\end{table}

\begin{table*}[t]
  \centering
  \small
  \setlength{\tabcolsep}{4pt}
  \begin{tabular}{lp{12.4cm}}
    \toprule
    \textbf{Part of speech} & \textbf{Template} \\
    \midrule
    noun (animate) & STEM -- \gt{n}\gt{an} -- (TONE) -- NUMBER (\gt{sg} $|$ \dng{-му}) -- (GENITIVE \dng{-ди} $|$ LOCATIVE \dng{-ни}/\dng{-шон} $|$ \dng{-дини}/\dng{-ди}/\dng{-ли}/\dng{-ни}) \\
    noun (inanimate) & STEM -- \gt{n}\gt{nn} -- (TONE) -- (same terminal slot; no number) \\
    proper noun & STEM -- \gt{np} -- (\dng{-жын}) -- (\dng{-му}) -- (\dng{-ди}) \\
    verb & STEM -- \gt{vblex} -- (TONE) -- (ASPECT \dng{-ли} $|$ \dng{-дини}/\dng{-ди} $|$ \dng{-ни} $|$ \dng{-гуә}) \\
    adjective & STEM -- \gt{adj} -- (TONE) -- (DEGREE \dng{-щер}/\dng{-дихын} $|$ PROGRESSIVE \dng{-дини}/\dng{-ди}) \\
    pronoun (personal) & STEM (incl.\ plural forms) -- \gt{prn}\gt{pers} -- (POSSESSIVE \dng{-ди}) \\
    \bottomrule
  \end{tabular}
  \caption{Morphotactic template: linear order of marker attachment (optional positions in parentheses, alternatives separated by $|$). The tone tag is emitted after the part-of-speech and animacy tags and before inflection. Note that the aspect slot on the verb is optional in the transducer, so a bare \gt{vblex} form is generable even though aspect marking is obligatory in the language (Section~\ref{sec:morphology}); and that the adjective admits only the progressive, not the full aspect set.}
  \label{tab:template}
\end{table*}

\section{Modelling Decisions with Page-Level Provenance}
\label{app:decisions}

Tables~\ref{tab:decisions} and~\ref{tab:decisions2} gather every modelled phenomenon into a single decision table: the phenomenon, the source(s) with page-level anchors, and how it is formalized in the analyzer. Page anchors are given for editions whose scans were read directly; where no page number can be substantiated (e.g.\ the author's web version of \citealp{salmi1984} does not preserve the journal pagination) the pages column shows a dash rather than a guess; pages known only through citation in another work are marked ``ap.'' (apud). The German-language \citet{dragunow1936} is cited by its own pagination (S.), the manuscript \citet{salmi2023} by draft pages (ms.).

\begin{table*}[p]
  \centering
  \footnotesize
  \setlength{\tabcolsep}{4pt}
  \begin{tabular}{p{4.3cm}p{4.6cm}p{7.6cm}}
    \toprule
    \textbf{Phenomenon} & \textbf{Source(s), pages} & \textbf{Formalization} \\
    \midrule
    \multicolumn{3}{l}{\emph{General}} \\
    \midrule
    Dungan an independent language; Gansu dialect as literary standard & \citealp{dragunow1936}, S.~34--35 & Lexicon extracted from a Gansu dictionary \citep{yanshansin2009}; Gansu is the lexical base and three-tone standard of the model, Shaanxi a separate hook lexicon. \\
    Gansu as the literary norm (first school teachers from Frunze) & \citealp{salmi2023}, ms.~p.~12 & Same architectural decision (Gansu as base lexicon), not a separate tag. \\
    Inventory of 14 parts of speech & \citealp{imazov1979}, pp.~74--75 & 11 of Imazov's 14 realized as POS tags (Table~\ref{tab:pos}); participle, converb and preposition are not tagged (\dng{ба} \hnz{把} is \gt{part}, \dng{лян} \hnz{連} is \gt{cnjcoo}); \gt{np} and \gt{cop} are added. \\
    More morphological markers than in Chinese & \citealp{salmi2023}, ms.~p.~27; \citealp{salmi1984},~--- & General observation motivating finite-state modelling; no dedicated tag. \\
    \midrule
    \multicolumn{3}{l}{\emph{Nominal morphology}} \\
    \midrule
    Criteria for the noun; animacy (\dng{сый}/\dng{са} test) & \citealp{imazov1979}, pp.~78--83 & Tag \gt{n} + feature \gt{an}/\gt{nn} (Table~\ref{tab:feats}); animacy gates the plural \dng{-му}. \\
    Animate plural \dng{-му}; genitive \dng{-ди} & \citealp{imazov1979}, pp.~78--83 & \gt{pl} on \dng{-му} (\dng{жын} $\to$ \dng{жынму}); \gt{gen} on \dng{-ди}. Analytic number (\dng{ну}, \dng{хошо}, \dng{хиге}) is syntactic, unmarked. \\
    Substantive suffixes \dng{-зы}/\dng{-р} (lexicalized, not segmented) & \citealp{tsunvazo1955}, pp.~75, 82--83; \citealp{dragunow1936}, S.~44 & Stored as whole dictionary stems (\dng{до} $\to$ \dng{дозы}); see \S\ref{sec:morphology} and Appendix~\ref{app:extnotes}. \\
    \dng{-зы}-marking correlates with stem tone & \citealp{polivanov1937}, pp.~41--58 & Treated as an explanation of lexicalization choice, not a separate rule; see Appendix~\ref{app:extnotes}. \\
    Classifier \dng{-гә} a suffix, not a counting word & \citealp{salmi2023}, ms.~p.~46; \citealp{imazov1982}, p.~66 (ap.\ Salmi 2023) & Tag \gt{cls}; 8 lexicon units (\dng{гә}/\dng{ба}/\dng{җон}/\dng{тё}/\dng{җы}/\dng{пи}/\dng{бын}/\dng{фу}), 5 productive; numeral+classifier a single complex (\dng{йи} + \hnz{個} $\to$ \dng{йигә}). \\
    \midrule
    \multicolumn{3}{l}{\emph{Verbal morphology (aspect--tense system)}} \\
    \midrule
    Obligatory aspect marking: bare verb makes an incomplete utterance & \citealp{dragunow1936}, S.~48; \citealp{salmi1984},~--- & The five markers of Table~\ref{tab:aspect} are the \gt{vblex} continuation classes. NB the transducer leaves the slot optional, so a bare \gt{vblex} form is generable; obligatoriness is not enforced. \\
    Perfective \dng{-ли} as ``demarcation point'' (aspect, not tense) & \citealp{dragunov1940}, pp.~15, 26 & Tag \gt{pst} a practical label; aspectual analysis and examples in \S\ref{sec:morphology} and Appendix~\ref{app:extnotes}. \\
    Inceptive reading of \dng{-ли} with statives (`came to know', `became cold') & \citealp{dragunov1940}, pp.~15--17, 26 & Same \gt{pst}; the inceptive reading is a contextual consequence of stem semantics, not a separate tag. \\
    Double \dng{-ли} with numeral objects (perfect of persistent situation) & \citealp{dragunov1940}, p.~32; \citealp{salmi2023}, ms.~\S6.3 & Formalized in both loci at once: both verb and object noun admit \gt{pst}/aspect clitics (\S\ref{sec:morphology}, analytic constructions). \\
    Imperfective/future \dng{-ни}: conditioned predicate; restriction with modals & \citealp{dragunov1940}, pp.~39--40, 44 & Tag \gt{fut} on \dng{-ни}; the modal restriction among the optionality cases (next row). \\
    Progressive \dng{-дини} vs stative \dng{-ди} contrast (background vs point) & \citealp{dragunov1940}, p.~26 & Both markers receive \gt{prog} (not distinguished by tag); cf.\ the divergence from Salmi's five-way analysis (Appendix~\ref{app:extnotes}). \\
    Experiential \dng{-гуә} (\hnz{過}, `at least once') & \citealp{salmi1984}, ---; \citealp{kalimov1968}, ---; \citealp{imazov1982},~--- & Tag \gt{exp} on \dng{-гуә} (\dng{нянгуә}). \\
    Aspect-marker loss under preposed negation (\dng{мә}/\dng{-ли}, \dng{бу}/\dng{-ни}) & \citealp{dragunov1940}, p.~58; \citealp{tsunvazoneg1949}, pp.~101--104 & NOT formalized at the morphological level~--- syntactic, requires context beyond the stem; see Section~\ref{sec:concl}. \\
    Aspect optionality (modals, statives, imperative, nominal predicate) & \citealp{dragunov1940}, p.~44 & Not blocked by a dedicated tag; control rests with the lexical entry of each stem. \\
    Five-way aspect system (different partition of categories) & \citealp{salmi1984},~--- & Divergence discussed in Appendix~\ref{app:extnotes} (progressive/stative as two tags instead of Salmi's imperfective). \\
    Past habitual \dng{-лэ}/\dng{-дилэ} (not modelled) & \citealp{salmi1984},~--- (crediting Dragunov 1952) & NOT in the lexicon~--- nearly unattested in the development corpus; deferred (Section~\ref{sec:concl}). \\
    \bottomrule
  \end{tabular}
  \caption{Modelling decisions with page-level provenance (part 1 of 2). A dash in the pages column means the edition has no citable pagination (web version), not a missing source.}
  \label{tab:decisions}
\end{table*}

\begin{table*}[p]
  \centering
  \scriptsize
  \setlength{\tabcolsep}{3.5pt}
  \renewcommand{\arraystretch}{0.96}
  \begin{tabular}{p{4.3cm}p{4.6cm}p{6.6cm}}
    \toprule
    \textbf{Phenomenon} & \textbf{Source(s), pages} & \textbf{Formalization} \\
    \midrule
    \multicolumn{3}{l}{\emph{Adjective}} \\
    \midrule
    Adjective as predicator; copular \dng{-сы} a copula suffix from \hnz{是} & \citealp{dragunow1936}, S.~42, 44; \citealp{zevakhina2001}, pp.~72--73, 75--76 & \gt{adj} admits aspect clitics (predicator status); the copula is a separate tag \gt{cop} (\dng{сы}/\dng{шы}, Table~\ref{tab:pos}). \\
    Predicative \dng{-ди} on the adjective & \citealp{imazov1979}, pp.~78--83; \citealp{zevakhina2001}, pp.~72--73, 75--76 & The same marker \dng{-ди} as the nominal genitive, in predicative-attributive role on \gt{adj} (marker polyfunctionality). \\
    Degrees of comparison \dng{-щер}/\dng{-дихын} & \citealp{imazov1982}, pp.~58, 60--61 (ap.\ Salmi 2023); \citealp{salmi2023}, ms.~p.~48 & \gt{comp} on \dng{-щер}, \gt{sup} on \dng{-дихын}; three competing readings of \dng{-дихын}: Appendix~\ref{app:extnotes}. \\
    Tonal criterion distinguishing adjective from verb & \citealp{dragunov1952},~--- (ap.\ Zevakhina 2001) & Not encoded as a tag; cited as an additional (for the model, non-morphological) argument for \gt{adj} as an autonomous class. \\
    \midrule
    \multicolumn{3}{l}{\emph{Tone and morphophonology}} \\
    \midrule
    Lexical tone: three tones in isolation; privative analysis & \citealp{polivanov1937}, pp.~41--58; \citealp{polivanov1937phon} & Tags \gt{t1}/\gt{t2}/\gt{t3} on $\approx$7{,}100 stems ($\approx$91\% of the lexicon); the privative reading of tone~I: Appendix~\ref{app:extnotes}. \\
    Three tones in isolation / four in connected speech; sandhi ``first $\to$ second'' & \citealp{salmi2023}, ms.~pp.~11--12 (crediting Zavyalova 1979) & NOT modelled (sandhi requires connected-speech context); see Appendix~\ref{app:extnotes}. \\
    Tone sandhi under reduplication (not modelled) & \citealp{tsunvazored1949}, pp.~67--70; \citealp{zavyalova1996}, pp.~64--65 & NOT modelled~--- reduplication as a whole outside this version; see also Appendix~\ref{app:extnotes}. \\
    \midrule
    \multicolumn{3}{l}{\emph{Other}} \\
    \midrule
    Full and short forms (elision of /i/, /\textschwa/); relevant to tokenization & \citealp{honkasalo2024}, \S3.2; \citealp{dragunow1936}, S.~38 (ap.\ Honkasalo) & Handled at lexicon-extraction level (tokenization), not by a grammatical tag. \\
    Noun locatives \dng{-ни} (\hnz{裡} `in') and \dng{-шон} (\hnz{上} `on') & \citealp{imazov1979}, pp.~78--83; \citealp{imazov1982} & \gt{loc} on \dng{-ни} (\dng{гонзы-ни} `in the bucket', \dng{фули-ни} `in the forest') and \dng{-шон} (\dng{дезы-шон} `on the plate'). \\
    Genitive \dng{-ди} polyfunctional (attributive, substantivising, adverbial; \dng{гўр-ди} `with a sharp sound') & \citealp{imazov1982}; \citealp{tsunvazo1963} & The one marker appears in several roles in the model: nominal genitive, predicative-attributive on adjectives (motivates multi-role \dng{-ди}). \\
    Noun admits aspect clitics (predicative marking of the V--O group) & \citealp{dragunov1940}; \citealp{zevakhina2001} & \gt{n} continuation classes include the aspect clitics \dng{-дини}/\dng{-ди}/\dng{-ни} (\S\ref{sec:morphology}). \\
    Collective ethnonyms \dng{грек-жын-му} `Greeks' (lit.\ `Greek-person-PL') & corpus-attested; provisional & Modelled as \dng{жын} \hnz{人} + \dng{-му} attachment to ethnonym stems; explicitly flagged as a \emph{provisional} generalization. \\
    Pronoun paradigm: person/number \dng{вә, ни, та, вәму, ниму, таму}; inclusive \dng{за}/\dng{заму} (\hnz{咱/咱們}) & \citealp{imazov1979} & \gt{prn} with subtypes \gt{pers} (person/number features) and \gt{incl}; pronouns take no preposed modifiers. \\
    Possessive pronoun forms in \dng{-ди} (\dng{вәму-ди} `our', \dng{ниму-ди} `your', \dng{таму-ди} `their') & \citealp{imazov1979} & \gt{px} slot in the pronominal template (Table~\ref{tab:template}). \\
    Demonstratives \dng{җыгә}/\dng{җәгә} (\hnz{這個}), \dng{нагә}/\dng{ныйгә} (\hnz{那個}); interrogatives \dng{са} (\hnz{啥}), \dng{сый} (\hnz{誰}), \dng{зуа}, \dng{зали} & \citealp{imazov1979} & \gt{dem} and \gt{itg} subtypes of \gt{prn}. \\
    Copula \dng{сы} (Gansu) / \dng{шы} (Shaanxi) marking the nominal predicate & \citealp{imazov1979} & Separate tag \gt{cop}, dialect hook lexicon. \\
    Postpositions: a closed class (\dng{литу} `inside', \dng{вэту} `outside', \dng{туни} `in front', \dng{хуту} `behind', \dng{диха} `below'); may take the genitive (\dng{либян} $\to$ \dng{либян-ди}) & \citealp{imazov1979} & \gt{post} closed class in the lexicon; genitive continuation enabled. \\
    Particles: a closed modal-focus class (\dng{ба} \hnz{把}, interrogative \dng{ма}, \dng{ла}, \dots) & \citealp{imazov1979} & \gt{part} closed class. \\
    Conjunctions (coordinating/subordinating) and interjections & \citealp{imazov1979} & \gt{cnjcoo}/\gt{cnjsub} and \gt{ij} closed classes. \\
    Function-word richness compensating minimal inflection & \citealp{imazov1979} & Motivates the closed function-word lexica above; no dedicated tag. \\
    Dialect differences (copula, personal pronouns, classifiers) & \citealp{salmi2023}, ms.~p.~13 & Realized as dialect hook lexicons (Table~\ref{tab:dialects}: \dng{сы}/\dng{шы}, \dng{вә}/\dng{ңә}, \dng{вәму}/\dng{ңәму}--\dng{заму}, \dng{таму}/\dng{ана}--\dng{тана}, \dng{гә}/\dng{гуә}--\dng{гы}). \\
    Morphological formation of adverbs & \citealp{tsunvazo1963}, pp.~1--2 of scan & Adverbs stored as whole dictionary stems with \gt{adv}; no productive adverbial suffix in the model (unlike Chinese). \\
    Reduplication as a productive mechanism (not modelled) & \citealp{tsunvazored1949}, pp.~67--70; \citealp{zevakhina2018},~--- & NOT modelled~--- outside the current lexicon version; see Section~\ref{sec:concl}. \\
    \bottomrule
  \end{tabular}
  \caption{Modelling decisions with page-level provenance (part 2 of 2).}
  \label{tab:decisions2}
\end{table*}

\section{Verbatim Source Quotations for the Modelled Rules}
\label{app:quotes}

For each modelled rule this appendix gives the source, page and paragraph (paragraphs are counted from the top of the page; the opening words of the paragraph are given in parentheses for verification), followed by a verbatim quotation. For rules supported by several sources, several quotations are given. Quotations keep the orthography and transcription of the originals (\citealp{dragunov1940} uses Latin transcription for the markers \mbox{-li/-ni/-dini}; \citealp{imazov1979} Cyrillic \dng{-ли}/\dng{-ни}/\dng{-дини}).

\paragraph{POS annotation: 14 parts of speech.}
\citealp{imazov1979}, p.~78, para.~6 ({\guillemotleft}Таким образом, разбираемые\dots{\guillemotright}): {\guillemotleft}…разбираемые слова обладают рядом семантических, синтаксических и морфологических признаков (значение предметности, функция подлежащего или дополнения…наличие категории числа и категории одушевлённости~--- неодушевлённости), которые отличают их от слов других классов. А такие слова…обычно относят к существительным.{\guillemotright}

\paragraph{Animate plural \dng{-му} \gt{pl} (only with nouns denoting persons).}
\citealp{imazov1979}, p.~78, para.~2 ({\guillemotleft}Отдельные из разбираемых\dots{\guillemotright}): {\guillemotleft}…морфему \dng{-му}, которая выражает значение множественности, например: \dng{хәйуан} (учитель)~--- \dng{хәйуанму} (учителя)…Правда, такой показатель множественности могут иметь лишь те немногие из них, которые обозначают лица.{\guillemotright}

\paragraph{Animacy \gt{an}/\gt{nn} (the \dng{сый?}/\dng{са?} test).}
\citealp{imazov1979}, p.~78, para.~5 ({\guillemotleft}…в дунганском\dots{\guillemotright}): {\guillemotleft}…в дунганском же различение одушевлённых и неодушевлённых имён существительных получило своё грамматическое выражение в постановке вопроса при установлении отнесённости слова к той или иной части речи.{\guillemotright} (the questions \dng{сый?} `who?' / \dng{са?} `what?')

\paragraph{Verbal aspect markers \dng{-дини}/\dng{-ни}/\dng{-ли}.}
\citealp{imazov1979}, p.~79, para.~5 ({\guillemotleft}Разбираемые слова имеют\dots{\guillemotright}): {\guillemotleft}Разбираемые слова имеют также характерные морфологические признаки~--- суффиксы \dng{-дини}, \dng{-ни}, \dng{-ли}, например: \dng{фадини} (играет), \dng{фани} (будет играть), \dng{фали} (играл); \dng{подини} (бежит), \dng{пони} (будет бежать), \dng{поли} (бежал)…{\guillemotright} \\
Same rule, second source: \citealp{dragunov1940}, p.~15, para.~2 ({\guillemotleft}Несмотря на кажущееся\dots{\guillemotright}): {\guillemotleft}…не требуют обязательного оформления на видовые предикативные суффиксы, т.~е.\ на суффиксы -dini, -li, -ni.{\guillemotright}

\paragraph{\dng{-ли} \gt{pst}~--- perfective, demarcation point (not a true past).}
\citealp{dragunov1940}, p.~15, \S I, para.\ ({\guillemotleft}Перфективное значение\dots{\guillemotright}): {\guillemotleft}Перфективное значение суффикса -li во всех этих примерах отчётливо выражено: то или иное действие или состояние имело место до такого-то определённого момента, после чего наступило новое действие или состояние. Переломный, демаркационный момент между концом прежнего (предыдущего) и началом нового (последующего) действия или состояния суффикс -li как раз именно и выражает.{\guillemotright}

\paragraph{Predicative / modal \dng{-ли} \gt{pst} on the nominal predicate (aspect clitic on the object noun in the V--O construction; marks completion of the measure, not of the action, which may continue).}
\citealp{dragunov1940}, p.~32, para.\ ({\guillemotleft}Конструкция с числительным\dots{\guillemotright}): {\guillemotleft}…отнюдь не к самому действию, а к дополнению с числительным определением. Действие же как таковое может и продолжаться{\guillemotright} (ibid.\ the example {\guillemotleft}к данному моменту я живу во Фрунзе уже три года{\guillemotright}; in our corpus~--- \dng{жын-ли} \hnz{人哩}, the hanzi gloss of the folk tale \hnz{把三個女子給哩人哩}).

\paragraph{Obligatory aspect marking (a bare verb is an unfinished word combination).}
\citealp{dragunow1936}, S.~48: {\guillemotleft}Insofern sie keine Suffixe haben, sind sie im Dunganischen \dots\ sinnlos und keineswegs als S\"atze zu betrachten, sondern als unvollendete Wortverbindungen. Hierin liegt einer der am sch\"arfsten ausgepr\"agten Unterschiede zwischen dem Dunganischen und dem Chinesischen.{\guillemotright}

\paragraph{\dng{-ни} \gt{fut}~--- prospective / conditioned (aspectual, not temporal).}
\citealp{dragunov1940}, p.~39, para.\ ({\guillemotleft}Случаи этого рода\dots{\guillemotright}): {\guillemotleft}Случаи этого рода отлично иллюстрируют различие в значении суффиксов -li и -ni. В каждом из приведённых примеров оба действия относятся к будущему, однако первое оформлено на суффикс -li, поскольку оно является сказуемым обусловливающим, второе же~--- на суффикс -ni, поскольку оно является сказуемым обусловленным.{\guillemotright}

\paragraph{\dng{-ни} is not an interrogative marker.}
\citealp{dragunov1940}, p.~39, para.\ ({\guillemotleft}Пример этот интересен\dots{\guillemotright}): {\guillemotleft}…приписывать дунганскому суффиксу -ni «вопросительное» значение, как это иногда делается по аналогии с китайским языком, было бы неверно…{\guillemotright}

\paragraph{Degrees of comparison \dng{-щер} \gt{comp}, \dng{-дихын} \gt{sup}.}
\citealp{imazov1979}, p.~80, para.~4 ({\guillemotleft}Последним присущи\dots{\guillemotright}): {\guillemotleft}…\dng{хи} (чёрный)~--- \dng{хищер} (чернее), \dng{го} (высокий)~--- \dng{гощер} (выше), \dng{бый} (белый)~--- \dng{быйдихн} (очень белый), \dng{ван} (мягкий)~--- \dng{вандихн} (очень мягкий)…Морфемы \dng{-щер} и \dng{-дихн}…являются отличительным морфологическим признаком…прилагательных.{\guillemotright}

\paragraph{\dng{-дихын} as intensifier of ``full predication'' (refinement of \gt{sup}).}
\citealp{zevakhina2001}, p.~75, para.~4 ({\guillemotleft}Первая из этих\dots{\guillemotright}): {\guillemotleft}…адъективный суффикс полной предикации, который в конструкциях такого вида является необязательным, но весьма желательным…{\guillemotright} (of the marker \dng{-дихын})

\paragraph{Predicative \dng{-ди} on the adjective with copula \dng{сы}.}
\citealp{imazov1979}, p.~80, para.~3 ({\guillemotleft}Слова данной категории\dots{\guillemotright}): {\guillemotleft}Слова данной категории в предложении являются частью составного именного сказуемого, если употребляются с суффиксом \dng{-ди}, а определяемый ими член предложения…употребляется со связкой \dng{сы}, например: \dng{Ванваньсы мутуди} (пиала деревянная)…Суффикс \dng{-ди} в данном случае определяет отнесённость…к прилагательным.{\guillemotright}

\paragraph{Adjective \gt{adj}~--- a predicator but an autonomous part of speech.}
\citealp{zevakhina2001}, p.~75, para.~6 ({\guillemotleft}Близость китайского\dots{\guillemotright}): {\guillemotleft}Близость китайского прилагательного по своим морфологическим свойствам к глаголам позволила ряду исследователей говорить об общей категории предикативов (ср.\ [Драгунов 1952: 12])…признать за прилагательными дунганского языка статус, хотя и особой, но всё же самостоятельной части речи.{\guillemotright}

\paragraph{Morphological criterion distinguishing adjective from verb~--- tone under reduplication.}
\citealp{dragunov1952} (apud \citealp{zevakhina2001}, p.~76, para.~1, {\guillemotleft}…различие отражается\dots{\guillemotright}): {\guillemotleft}Он отмечает, что их различие отражается в тональной структуре сложных слов, образованных путём удвоения качественной, с одной стороны, и глагольной, с другой стороны, морфем [Драгунов 1952: 168--170].{\guillemotright}

\paragraph{Polyfunctional \dng{-ди} (function element; with onomatopoeia).}
\citealp{tsunvazo1963}, p.~52, para.~2 ({\guillemotleft}В дунганском языке\dots{\guillemotright}): {\guillemotleft}В дунганском языке звукоподражательные слова сами по себе, ни в суффиксальном оформлении самостоятельно не употребляются. Почти всегда они используются в связанной речи, обычно в сочетании с суффиксом \dng{-ди}.{\guillemotright}

\paragraph{Adverbs stored as whole stems (no dedicated adverbial suffix).}
\citealp{tsunvazo1963}, p.~53, last para.\ ({\guillemotleft}Наречия, обозначающие\dots{\guillemotright}): {\guillemotleft}Наречия, обозначающие одни и те же понятия, в дунганском и китайском языках в ряде случаев имеют различное оформление (разные предлоги, суффиксы и знаменательные морфемы).{\guillemotright}

\paragraph{Counting words (classifiers) with the numeral.}
\citealp{imazov1979}, p.~81, para.~1 ({\guillemotleft}…количества в языке\dots{\guillemotright}): {\guillemotleft}…чаще всего сочетаются со счётными словами типа \dng{фон} (пара), \dng{ба} (горсть)…например: \dng{Ве мали лён-фон хэ} (я купил две пары обуви)…{\guillemotright} \\
Same rule, second source: \citealp{kalimov1958}, title of the work: {\guillemotleft}Счётные суффиксы в современном дунганском языке{\guillemotright} (a dedicated study of the counting markers).

\paragraph{Lexical tone \gt{t1}/\gt{t2}/\gt{t3} (Gansu: three tones, notation I/II/III).}
\citealp{polivanov1937}, p.~41, \S1, para.~1 ({\guillemotleft}Характерным отличием\dots{\guillemotright}): {\guillemotleft}Характерным отличием всех «тибето-китайских» языков является наличие так называемых «тонов», т.~е.\ различных мелодий голосового тона, присущих определённым слогам-морфемам…{\guillemotright}

\paragraph{Three tonemes, numbered I, II, III.}
\citealp{polivanov1937}, p.~56, \S9 (heading): {\guillemotleft}\S~9. Примеры дунганских односложных слов под разными (I, II, III) ГС тонами{\guillemotright} (followed by the rows \dng{tan} `field' (I) / `carpet' (II) / `coal' (III), etc.).

\paragraph{twol: deletion of the morpheme boundaries \texttt{\%>}, \texttt{\%+} $\to$ $\emptyset$.}
Engineering rule; grounding: \citealp{imazov1979}, p.~78, para.~6: markers attach to the stem without morphophonological changes (Dungan is analytic); the boundary is needed only for morphotactics and is erased on the surface ({\guillemotleft}…наличие у некоторых из них определённых флексий…{\guillemotright}, Imazov 1979, p.~78).

\paragraph{twol (documented, vacuous): \dng{ў}/\dng{у} after labials.}
\citealp{salmi2018}; Russian Wikipedia, {\guillemotleft}Дунганский язык{\guillemotright}; descriptive phonology: the rule is carried as a documented constraint and is vacuous in the model (stems are stored in final orthography). After the labials \dng{б п м ф в} the closed /u/ is written \dng{у} (\dng{бу} \hnz{不}, \dng{фу} \hnz{服}), not \dng{ў}. No verbatim quotation from the accessible Soviet sources; the rule is not active.

\paragraph{twol (documented, vacuous): Shaanxi \dng{дь}/\dng{ть} $\to$ \dng{җь}/\dng{чь}.}
\citealp{dragunow1936}; descriptive phonology: carried as a documented constraint, vacuous in the model (Shaanxi stems are stored already palatalized). The historical soft \dng{дь-}/\dng{ть-} yields Shaanxi \dng{җь-}/\dng{чь-}; at syllable level \dng{дьа}$\to$\dng{җьа}, \dng{тьи}$\to$\dng{чьи}. The rule is not active; no specific dictionary pair is cited, to avoid inaccuracy.

\section{Extended Notes: Derivation, Tone History and the Transducer Graph}
\label{app:extnotes}

\paragraph{Derivational suffixes \dng{-зы}/\dng{-р} (full discussion).}
A special case is derivational, not inflectional, suffixation. Most disyllabic Dungan nouns arise by compounding, in which the tone of one component often changes, and a number of monosyllabic roots become autonomous only when dressed with the ``substantive'' suffixes \dng{-зы} (\hnz{子}) or \dng{-р} (\hnz{兒}): \dng{до} `sword'~--- \dng{дозы} `knife', and the bound root \dng{кў-} (free only in compounds, cf.\ \dng{кўтуй} `trouser leg', where \dng{уй} is `leg')~--- \dng{кўзы} `trousers' \citep[p.~79]{tsunvazo1955}. The model treats such formations as ready-made dictionary units rather than productively detachable markers: \dng{-зы} is largely lexicalized in the modern language (cf.\ \dng{фонзы} `house', \dng{гонзы} `bucket', which have lost their diminutive value), and segmenting it productively would create spurious homonymy. Interestingly, the very choice of \dng{-зы}-marking correlates statistically with stem tone: \citet{polivanov1937} observes that stems under the longer tone (in Gansu, tone~I) more often remain monosyllabic, while stems under the short tones gravitate to \dng{-зы}-marking. Modelling derivation is therefore deferred.

\paragraph{Perfective \dng{-ли} and prospective \dng{-ни} (full discussion).}
The model's tag \gt{pst} (`past') is a practical label: \dng{-ли} is in essence an aspectual, not a temporal marker. \citet[p.~15]{dragunov1940} analyses it as a ``demarcation point''~--- the turning point between a previous and a new state. This explains why with atelic stems \dng{-ли} yields not a past but an inceptive reading: with the stative verbs \dng{зы} `know' and \dng{ю} `have' the meanings `came to know', `came to have' arise (\dng{юли} `(I) came to have'), and with a predicative adjective, a change of state (\dng{лын-ли} `it became cold'). The same accounts for the combination of \dng{-ли} with the negator \dng{бу}, which conveys cessation (`quit smoking') rather than a plain past. Symmetrically, the prospective \dng{-ни} is above all aspectual rather than temporal: \citet[pp.~39--40]{dragunov1940} notes that in conditional and temporal constructions, even when both events are in the future, the conditioning (prior) predicate takes perfective \dng{-ли} and the conditioned (subsequent) one takes prospective \dng{-ни} (`when I arrive [\dng{-ли}], I will speak [\dng{-ни}]'); the \dng{-ли}/\dng{-ни} opposition contrasts completion and prospection, not past and future. Hence also the `not yet' construction (\dng{хан мә V-ни}), where the negator \dng{мә} and prospective \dng{-ни} regularly co-occur: the event has not yet arrived but is in prospect. The obligatoriness of aspect marking has a known list of exceptions (negated verb, modal verbs, statives, the imperative, nominal and temporal-age predicates), in which the marker is optional \citep[p.~44]{dragunov1940}. The systematic correspondence of positive and negative forms, in which a preposed negative particle triggers regular loss of the aspect marker, was described by \citet[p.~58]{dragunov1940}; see Section~\ref{sec:concl}.

\paragraph{Relation to Salmi's analysis (full discussion).}
The five-member system is close to Salmi's independent analysis, which also distinguishes five aspect--tense categories for Central Asian Dungan, but with a different partition: past habitual \dng{-лэ}/\dng{-дилэ}, experiential perfect, imperfective, perfective and future \citep{salmi1984}. The divergence is twofold. First, our model keeps progressive \dng{-дини} and stative \dng{-ди} as separate markers, whereas \citet{salmi1984} unites them into an imperfective (non-final form \dng{-ди}, final \dng{-дини}). Second, the past habitual he describes (earlier also \citealp{dragunov1952})~--- \dng{-лэ} in stative clauses (with the copula, with stative adjectives, with \dng{мәю} `not have') and \dng{-дилэ} in actional ones~--- is \emph{not} formalized in the present model: it is nearly unattested in our corpus of written Dungan and is deferred. The stative/actional split itself, with which \citet{salmi1984} explains the defectiveness of the aspect paradigm in stative contexts, agrees with the optionality contexts listed in the preceding discussion of \dng{-ли}/\dng{-ни}.

\paragraph{Degrees of comparison (full discussion).}
The adjective carries \gt{adj} and admits the comparative \dng{-щер} (Imazov's examples: \dng{хи} `black'~--- \dng{хищер} `blacker', \dng{го} `tall'~--- \dng{гощер} `taller'), the superlative (elative) \dng{-дихын} (\hnz{得很}: \dng{бый} `white'~--- \dng{быйдихын} `very white') \citep{imazov1979,imazov1982}, and predicative use with aspect particles; by \citet{zevakhina2001} the predicative adjective takes the whole verbal aspect set (\dng{-ли}, \dng{-ни}, \dng{-дини}, cf.\ \dng{Ни хо-дини ма?} `How do you do?', lit.\ `are you in a good state?'), though \dng{-ли}/\dng{-ни} on adjectives are nearly unattested in our corpus and not modelled productively. \dng{-дихын} is described both as a superlative \citep{imazov1982} and~--- matching its etymology as the intensifier \hnz{得很} `very'~--- as an optional ``suffix of full predication'' \citep{zevakhina2001}, cf.\ \dng{Кў-зы та-шон зый-дихын} `the trousers are (too) tight for him'; all descriptions agree on the intensive-predicative reading, and the model adopts the superlative tag.

\paragraph{Lexical tone: orthographic history and deplacement.}
Introducing tone tags can be seen as restoring~--- at the level of grammatical annotation rather than orthography~--- the distinctive feature that the 1928--1932 Latin script could in principle have conveyed~--- but did not, being aligned with Latinxua Sin Wenz, which programmatically leaves tone unwritten. Tellingly, leaving tone unwritten was a deliberate decision: \citet{polivanov1937} himself argued against \emph{obligatory} tone marking, holding that it would encumber Dungan writing as obligatory stress marks would encumber Russian (cf.\ \dng{з\'{а}мок} `castle' / \dng{зам\'{о}к} `lock'), and allowed only an optional mark in ambiguous cases and in pedagogical literature. The homography our tags resolve is thus built into the very norm of the script.

Nor is the incompleteness of the orthographic record exhausted by that. \citet{polivanov1937} analyses the three-tone system \emph{privatively}: tone~I is ``null'' (level, no pitch movement), opposed to the ``positive'' tones II (falling) and III (rising). With this null character of tone~I he links the phenomenon he calls \emph{d\'eplacement} (депласация)~--- the tone-conditioned transfer of the expiratory stress \citep[p.~53]{polivanov1937}: in suffixal formations stress falls on the root by default, but if the root carries tone~I it shifts to the suffix (\dng{ма}\textsuperscript{1} `hemp' + \dng{-ди} $\to$ \dng{ма-д\'{и}}, but \dng{ма}\textsuperscript{2} `horse' + \dng{-ди} $\to$ \dng{м\'{а}-ди}). Since Cyrillic marks neither tone nor stress, pairs such as \dng{поли} `fled' (tone~I, suffix stress) and \dng{поли} `ran' (tone~II, root stress) are doubly indistinguishable in writing.

\paragraph{Invertibility examples and tag order.}
The tonal disambiguation and round-trip of \dng{задё}:

\begin{center}
\small
\begin{tabular}{l}
(3) \dng{задё} $\to$ \dng{задё}\gt{vblex}\gt{t1} `suck out'; \\
\hphantom{(3) }\dng{задё}\gt{vblex}\gt{t2} `fence off'; \\
\hphantom{(3) }\dng{задё}\gt{vblex}\gt{t3} `blow up' \\[2pt]
(4) \dng{задё}\gt{vblex}\gt{t3}\gt{pst} $\leftrightarrow$ \dng{задёли} `blew up' \\
\end{tabular}
\end{center}

The form \dng{жынму} is analysed as \dng{жын}\gt{n}\gt{an}{\gt{t1}}\gt{pl}, and the same tag string generates exactly \dng{жынму}:

\begin{center}
\small
\setlength{\tabcolsep}{2.5pt}%
\begin{tabular}{llll}
(5) & \dng{жын-му} & \dng{жын}\gt{n}\gt{an}\gt{t1}\gt{pl} & `people' \\
(6) & \dng{задё-ли} & \dng{задё}\gt{vblex}\gt{t3}\gt{pst} & `blew up' \\
(7) & \dng{самария-ди} & \dng{самария}\gt{np}\gt{gen} & `of Samaria' \\
\end{tabular}
\end{center}

Tag order follows Apertium and GiellaLT practice: part of speech first, then the lexical tone feature, then inflectional features (number, aspect). This order is an engineering convention of the annotation, not a claim about the structure of the Dungan word; the lexically inherent feature (tone) sits closer to the stem than the inflectional ones.

\begin{figure}[t]
  \centering
  \includegraphics[width=0.98\columnwidth]{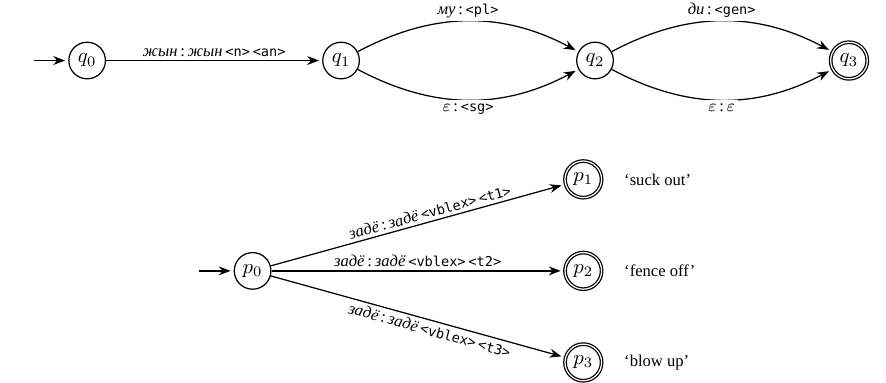}
  \caption{Two transducer fragments. Top: nominal morphotactics (the path \dng{жынму} $\leftrightarrow$ \dng{жын}\gt{n}\gt{an}\gt{t1}\gt{pl}); bottom: resolution of tonal homography of the stem \dng{задё} (one surface spelling~--- three analyses differing only in the tone tag). Arcs are labelled ``lower level : upper level''; $\varepsilon$ is the empty transition.}
  \label{fig:fst}
\end{figure}

\paragraph{The transducer as a graph.}
Figure~\ref{fig:fst} shows two fragments of the transducer. The upper fragment illustrates nominal morphotactics: the stem \dng{жын} takes the automaton from the initial state to a state from which optional number and genitive slots lead to an accepting state; each arc is labelled with a ``surface segment : upper-level tag'' pair, so the path \dng{жынму} reads on the lower level as the surface form and on the upper level as the analysis \dng{жын}\gt{n}\gt{an}\gt{t1}\gt{pl}. The empty transition ($\varepsilon$) corresponds to the unmarked value (singular; no genitive). The lower fragment illustrates tonal disambiguation: one and the same surface string \dng{задё} maps to three analyses differing only in the tone tag~--- it is these parallel arcs that restore the distinction the orthography does not record.

\section{A Survey of Dungan Language Resources}
\label{app:resources}

Tables~\ref{tab:res1}--\ref{tab:res4} give an annotated survey of 70 resources on Dungan and adjacent topics, grouped by type. Availability: \emph{open} = freely accessible electronic version; \emph{part.} = accessible with caveats (registration, partial, cache only); \emph{print} = print only. Titles are kept in their original language. The Oslo Digital Archive of Dungan Studies (ODADS, row~53) hosts scans of most of the classic literature.

\begin{table*}[p]
  \centering
  \footnotesize
  \setlength{\tabcolsep}{3pt}
  \begin{tabular}{rp{2.5cm}p{4.3cm}p{1.7cm}p{6.0cm}}
    \toprule
    \textbf{\#} & \textbf{Author} & \textbf{Title} & \textbf{Year / avail.} & \textbf{Source (URL)} \\
    \midrule
    \multicolumn{5}{l}{\emph{Encyclopaedias and reference works (11)}} \\
    \midrule
    1 & Калимов А. (ред.\ В.~Н. Ярцева) & Дунганский язык (статья в ЛЭС) & 1990, open & \url{https://tapemark.narod.ru/les/145b.html} \\
    2 &~--- & Дунганский язык (Энциклопедия Кругосвет) & ---, open & \url{https://www.krugosvet.ru/enc/gumanitarnye_nauki/lingvistika/DUNGANSKI_YAZIK.html} \\
    3 &~--- & Дунганский язык (Большая российская энциклопедия) & ---, open & \url{https://old.bigenc.ru/linguistics/text/5229551} \\
    4 &~--- & Дунганский язык (Русская Википедия) & ---, open & \url{https://ru.wikipedia.org/wiki/Дунганский_язык} \\
    5 &~--- & Дунганский язык (RuWiki) & ---, open & \url{https://ru.ruwiki.ru/wiki/Дунганский_язык} \\
    6 &~--- & Дунганский язык (Традиция / traditio.wiki) & ---, open & \url{https://traditio.wiki/Дунганский_язык} \\
    7 &~--- & Дунганский язык (портал {\guillemotleft}Языки народов России{\guillemotright}, МПГУ) & ---, open & \url{https://mpgu.su/ob-mpgu/struktura/biblioteka/kulturno-prosvetitelskie-proektyi/jazyki-narodov-rossii/dunganskij-jazyk/} \\
    8 &~--- & Дунганский (портал {\guillemotleft}Малые языки России{\guillemotright}, ИЛИ РАН) & ---, open & \url{https://minlang.iling-ran.ru/lang/dunganskiy} \\
    9 & Max Planck Institute (Hammarström H. et al.) & Glottolog 5.2: Dungan (dung1253) & 2025 (v.~5.2), open & \url{https://glottolog.org/resource/languoid/id/dung1253} \\
    10 & SIL International & Ethnologue: Dungan (dng) & 2018+ (annual), part. & \url{https://www.ethnologue.com/language/dng} \\
    11 &~--- & Dungan language (English Wikipedia) & ---, open & \url{https://en.wikipedia.org/wiki/Dungan_language} \\
    \midrule
    \multicolumn{5}{l}{\emph{Dictionaries (5)}} \\
    \midrule
    12 & Яншансин Ю. & Краткий дунганско-русский словарь & 2009 (2nd ed.), open & \url{https://ibtrussia.org/sites/default/files/pdf/Dungan_Dictionary.pdf} \\
    13 & Сушанло М.~Я. (ред.) & Краткий дунганско-русский словарь (Җеёди хуэйзў-вурус хуадян) & 1968, print &~--- \\
    14 &~--- & Русско-дунганский словарь, тт.~1--3 & 1981, print &~--- \\
    15 & Salmi O. & Dungan-English Dictionary & 2018, part. & \url{http://www.tsalo.fi/Dungan-English%20Dictionary.pdf} \\
    16 & Шинло Л.~Т. (сост.) & Приложение. Термины родства у дунган фрунзенской, александровской и ырдыкской групп & 1965, open & \url{https://www.kromatikon.eu/odads/Library/Dictionaries/Prilozhenie.%20Terminy%20rodstva%20_%201965.PDF} \\
    \bottomrule
  \end{tabular}
  \caption{Dungan resource survey, part 1: reference works and dictionaries.}
  \label{tab:res1}
\end{table*}

\begin{table*}[p]
  \centering
  \footnotesize
  \setlength{\tabcolsep}{3pt}
  \begin{tabular}{rp{2.5cm}p{4.3cm}p{1.7cm}p{6.0cm}}
    \toprule
    \textbf{\#} & \textbf{Author} & \textbf{Title} & \textbf{Year / avail.} & \textbf{Source (URL)} \\
    \midrule
    \multicolumn{5}{l}{\emph{Grammar: primary literature (36)~--- part 1}} \\
    \midrule
    17 & Драгунов А.~А., Драгунова Е.~Н. & Дунганский язык & 1937, open & \url{https://www.academia.edu/121980989/} \\
    18 & Драгунов А.~А. & Исследования в области дунганской грамматики. Ч.~1: Категория вида и времени в дунганском языке (диалект Ганьсу) & 1940, open & \url{https://www.kromatikon.eu/odads/Library/Studies_Language/Dragunov%201940%20-%20Issledovanija%20Dungan%20Aspect%20Time.pdf} \\
    19 & Dragunov A., Dragunova E. & Über die dunganische Sprache & 1936, open & \url{https://www.kromatikon.eu/odads/Library/Studies_Language/Dragunow%201936%20-%20Ueber%20die%20dunganische%20Sprache.PDF} \\
    20 & Поливанов Е.~Д. & Музыкальное слогоударение, или {\guillemotleft}тоны{\guillemotright} дунганского языка & 1937, open & \url{https://www.kromatikon.eu/odads/Library/Studies_Language/Polivanov%201937%20-%20Muzykal%27noe%20slogoudarenie.pdf} \\
    21 & Поливанов Е.~Д. & Фонологическая система ганьсуйского наречия дунганского языка & 1937, print &~--- \\
    22 & Калимов А. & Дунганский язык & 1968, open & \url{https://litgu.ru/knigi/guman_nauki/426979-serija-jazyki-narodov-sssr-5-tomov.html} \\
    23 & Калимов А. & Счётные суффиксы в современном дунганском языке & 1958, print &~--- \\
    24 & Калимов А. & Грамматические особенности счётных слов, счётных суффиксов и единиц измерения… & 1951, print &~--- \\
    25 & Калимов А. & Несколько замечаний о путях развития дунганского языка & 1975, open & \url{https://www.kromatikon.eu/odads/Library/Studies_Language/} \\
    26 & Калимов А. & А.~А. Драгунов~--- основоположник дунганского языкознания & 1980, open & \url{https://www.kromatikon.eu/odads/Library/Dungan_Studies/Kalimov%201980%20-%20A.%20A.%20Dragunov.pdf} \\
    27 & Имазов М.~Х. & Очерки по морфологии дунганского языка & 1982, print &~--- \\
    28 & Имазов М.~Х. & Очерки по синтаксису дунганского языка & 1987, print &~--- \\
    29 & Имазов М.~Х. & Грамматика дунганского языка & 1993, open & \url{https://www.twirpx.com/file/2705549} \\
    30 & Имазов М.~Х. & Грамматика дунганского языка (автореферат докт.\ дисс.) & 1994, open & \url{https://cheloveknauka.com/grammatika-dunganskogo-yazyka} \\
    31 & Имазов М.~Х. & Фонетика дунганского языка & 1975, print &~--- \\
    32 & Имазов М.~Х. & О частях речи в дунганском языке & 1979, open & \url{https://www.kromatikon.eu/odads/Library/Studies_Language/Imazov%201979%20-%20O%20chastjakh%20rechi.PDF} \\
    33 & Завьялова О.~И. & Диалекты Ганьсу & 1979, print &~--- \\
    34 & Завьялова О.~И. & Тоны в дунганском языке & 1973, print &~--- \\
    35 & Завьялова О.~И. & Тоны в шэньсийском диалекте дунганского языка & 1978, print &~--- \\
    36 & Завьялова О.~И. & Диалекты китайского языка & 1996, open & \url{https://www.klex.ru/1hex} \\
    \bottomrule
  \end{tabular}
  \caption{Dungan resource survey, part 2: primary grammatical literature (first half).}
  \label{tab:res2}
\end{table*}

\begin{table*}[p]
  \centering
  \footnotesize
  \setlength{\tabcolsep}{3pt}
  \begin{tabular}{rp{2.5cm}p{4.3cm}p{1.7cm}p{6.0cm}}
    \toprule
    \textbf{\#} & \textbf{Author} & \textbf{Title} & \textbf{Year / avail.} & \textbf{Source (URL)} \\
    \midrule
    \multicolumn{5}{l}{\emph{Grammar: primary literature~--- part 2}} \\
    \midrule
    37 & Зевахина Т.~С. & Функционально-грамматическая параметризация прилагательного (по данным полевого исследования дунганского языка) & 2001, open & \url{https://www.kromatikon.eu/odads/Library/Studies_Language/Zevakhina%202001%20-%20Funkcional%27no-grammaticheskaja%20parametrizacija.pdf} \\
    38 & Зевахина Т.~С., Имазов М.~Х. & Дунганский язык & 1997, print &~--- \\
    39 & Зевахина Т.~С. & О словарном описании многозначных прилагательных (дунганский) & 1984, print &~--- \\
    40 & Зевахина Т.~С. & О функционально-грамматических аспектах словарного описания дунганского прилагательного & 1986, print &~--- \\
    41 & Цунвазо Ю. & Повторение в дунганском языке & 1949, open & \url{https://www.kromatikon.eu/odads/Library/Studies_Language/Cunvazo%201949%20-%20Povtorenie%20v%20dunganskom.PDF} \\
    42 & Цунвазо Ю. & К вопросу о средствах выражения отрицания в дунганском языке & 1949, open & \url{https://www.kromatikon.eu/odads/Library/Studies_Language/Cunvazo%201949%20-%20K%20voprosu.PDF} \\
    43 & Цунвазо Ю. & К вопросу о способах словообразования в дунганском языке & 1955, open & \url{https://www.kromatikon.eu/odads/Library/Studies_Language/Cunvazo%201955%20-%20Slovoobrazovanie.PDF} \\
    44 & Цунвазо Ю. & О двусложных существительных в современном дунганском языке & 1956, open & \url{https://www.kromatikon.eu/odads/Library/Studies_Language/Cunvazo%201956%20-%20Compounds.PDF} \\
    45 & Цунвазо Ю. & О морфологическом способе словообразования наречий в дунганском языке & 1963, open & \url{https://www.kromatikon.eu/odads/Library/Studies_Language/Cunvazo%201963%20-%20O%20morfologicheskom%20sposobe.PDF} \\
    46 & Salmi O. & The Aspectual System of Soviet Dungan & 1984, open & \url{https://www.kromatikon.eu/odads/Library/Studies_Language/Salmi%201984%20-%20The%20Aspectual%20System%20of%20Soviet%20Dungan.pdf} \\
    47 & Salmi O. & Tone and Stress in Soviet Dungan & 1980, print &~--- \\
    48 & Salmi O. & The Syntax of Central Asian Dungan (ms., {\guillemotleft}Dungan Syntax{\guillemotright}) & c.~2023 (archived), open & \url{https://www.kromatikon.eu/odads/Library/Studies_Language/Salmi%20-%20Dungan%20Syntax.pdf} \\
    49 & Salmi O. & Notes on the Shaanxi Dialect of Central-Asian Dungan (ms.) & c.~2023 (archived), open & \url{https://www.kromatikon.eu/odads/Library/Studies_Language/Salmi%20-%20Notes%20on%20the%20Shaanxi%20Dialect%20of%20Central-Asian%20Dungan.pdf} \\
    50 & Hai Feng \hnz{海峰} & \hnz{中亞}\allowbreak \hnz{東干語}\allowbreak -му\allowbreak \hnz{「們」}\allowbreak \hnz{的使用特點} (on the use of \dng{-му} \hnz{們}) & 2004, open & \url{https://www.kromatikon.eu/odads/Library/Studies_Language/Hai%20Feng%202004%20-%20Zhongya%20Dongganyu%20-mu.pdf} \\
    51 & Wang Jingrong \hnz{王景榮} & \hnz{東干語、}\allowbreak \hnz{漢語}\allowbreak \hnz{烏魯木齊}\allowbreak \hnz{方言}\allowbreak \hnz{「完成」}\allowbreak \hnz{體貌助詞}\allowbreak \hnz{「哩/咧」} (the perfective particle \hnz{哩/咧}) & 2006, open & \url{https://www.kromatikon.eu/odads/Library/Studies_Language/Wang%20Jingrong%202006%20-%20Dongganyu,%20Hanyu.pdf} \\
    52 & Wang Jingrong \hnz{王景榮} & \hnz{形容詞}\allowbreak \hnz{後的}\allowbreak \hnz{助詞}\allowbreak \hnz{「下」}[x$\alpha$] (the particle \hnz{下} after adjectives) & 2004, open & \url{https://www.kromatikon.eu/odads/Library/Studies_Language/Wang%20Jingrong%202004%20-%20Dongganyu,%20Hanyu.pdf} \\
    \bottomrule
  \end{tabular}
  \caption{Dungan resource survey, part 3: primary grammatical literature (second half).}
  \label{tab:res3}
\end{table*}

\begin{table*}[p]
  \centering
  \footnotesize
  \setlength{\tabcolsep}{3pt}
  \begin{tabular}{rp{2.5cm}p{4.3cm}p{1.7cm}p{6.0cm}}
    \toprule
    \textbf{\#} & \textbf{Author} & \textbf{Title} & \textbf{Year / avail.} & \textbf{Source (URL)} \\
    \midrule
    \multicolumn{5}{l}{\emph{Corpora, databases, archives (8)}} \\
    \midrule
    53 & Spira I. (comp.) & Oslo Digital Archive of Dungan Studies (ODADS) / kromatikon & ongoing (last upd.\ 2023), open & \url{https://www.kromatikon.eu/odads/index.html} \\
    54 & Wichmann S., Holman E.~W., Brown C.~H. (eds.) & ASJP Database (Automated Similarity Judgment Program) & 2016, open & \url{https://asjp.clld.org/languages/DUNGAN.txt} \\
    55 & Рифтин Б.~Л. (ред.) & Дунганские народные сказки и предания & 1977, part. & \url{https://www.kromatikon.eu/odads/folktales.html} \\
    56 & (dng-fst project) & dungan\_parallel.json~--- parallel corpus & 2026, open & in the project archive \\
    57 & (dng-fst project) & dungan\_wikipedia.json~--- Shaanxi text corpus & 2026, open & in the project archive \\
    58 & Scannell K.~P. & Crúbadán: Dungan data (web corpus) & 2018 (snapshot); project 2007, archived & \url{https://github.com/kscanne/crubadan} \\
    59 & Open Language Archives Community & OLAC: Dungan resources (aggregator) & updated, open & \url{http://www.language-archives.org/language/dng} \\
    60 & LINGUIST List (eds.\ D.~Cavar, M.~E. Cavar) & LINGUIST List: Dungan resources & 2022, open & \url{https://linguistlist.org} \\
    \midrule
    \multicolumn{5}{l}{\emph{Methodological parallels (FST) (4)}} \\
    \midrule
    61 & Ahmadi S., Hassani H. & Towards Finite-State Morphology of Kurdish & 2020, open & \url{https://arxiv.org/abs/2005.10652} \\
    62 & Rahi R., Pushp S., Khan A., Sinha S.~K. & A Finite State Transducer Based Morphological Analyzer of Maithili Language & 2020, open & \url{https://arxiv.org/abs/2003.00234} \\
    63 & Naserzade M., Mahmudi A., Veisi H., Hosseini H., MohammadAmini M. & CKMorph: A Comprehensive Morphological Analyzer for Central Kurdish & 2023, open & \url{https://doi.org/10.1007/s42803-022-00062-7} \\
    64 & Apertium & apertium-kaz: morphological transducer and disambiguator for Kazakh & ongoing, open & \url{https://github.com/apertium/apertium-kaz} \\
    \midrule
    \multicolumn{5}{l}{\emph{Tools and technologies (3)}} \\
    \midrule
    65 & Lindén K., Axelson E., Hardwick S., Pirinen T.~A., Silfverberg M. & HFST~--- Framework for Compiling and Applying Morphologies & 2011, open & \url{https://link.springer.com/chapter/10.1007/978-3-642-23138-4_5} \\
    66 &~--- & Apertium (morphological transducers / MT platform) & ---, open & \url{https://www.apertium.org} \\
    67 &~--- & GiellaLT (FST infrastructure for smaller languages) & ---, open & \url{https://giellalt.uit.no} \\
    \midrule
    \multicolumn{5}{l}{\emph{Secondary / background (3)}} \\
    \midrule
    68 &~--- & Негативные конструкции в дунганском языке (статья, КиберЛенинка) & ---, open & \url{https://cyberleninka.ru/article/n/negativnye-konstruktsii-v-dunganskom-yazyke-s-tochki-zreniya-tipologicheskoy-perspektivy-ego-razvitiya-1} \\
    69 &~--- & Большой мир китайского языка (изд.\ 2-е) & ---, open & \url{https://www.academia.edu/42660534/} \\
    70 & Rimsky-Korsakoff Dyer S. & Soviet Dungan Kolkhozes in the Kirghiz SSR and the Kazakh SSR & 1979, print &~--- \\
    \bottomrule
  \end{tabular}
  \caption{Dungan resource survey, part 4: corpora and archives, FST parallels, tools, background.}
  \label{tab:res4}
\end{table*}

\end{document}